\documentclass[10pt,journal,compsoc]{IEEEtran}
\usepackage{amssymb,amsmath,amsthm}
\usepackage{graphicx}
\usepackage{subfigure}
\usepackage{capt-of}
\usepackage{multirow}
\usepackage{bbding}

\usepackage{tikz}
\usepackage{comment}
\usepackage[utf8]{inputenc}
\usepackage[english]{babel}
\usepackage{amssymb,amsmath,amsthm} 
\usepackage{color}
\usepackage{wrapfig,lipsum,booktabs}

\usepackage{xspace}
\makeatletter
\DeclareRobustCommand\onedot{\futurelet\@let@token\@onedot}
\def\@onedot{\ifx\@let@token.\else.\null\fi\xspace}
\usepackage{framed}
\usepackage{graphicx}
\usepackage{booktabs}
\usepackage{url}
\usepackage{paralist}
\usepackage[accsupp]{axessibility}
\usepackage[pagebackref,breaklinks,colorlinks]{hyperref}
\newsavebox\CBox
\def\textBF#1{\sbox\CBox{#1}\resizebox{\wd\CBox}{\ht\CBox}{\textbf{#1}}}

\def\eg{\emph{e.g}\onedot} 
\def\ie{\emph{i.e}\onedot} 
 
\def\etc{\emph{etc}\onedot}

\def\etal{\emph{et al}\onedot}

\def\REV#1{{{#1}}}

\def\REVX#1{{{#1}}}

\ifCLASSOPTIONcompsoc
  \usepackage[nocompress]{cite}
\else
  \usepackage{cite}
\fi
\ifCLASSINFOpdf
\else
\fi
\usepackage{graphicx}

\usepackage{url}

\begin{document}
%
\title{Deep Non-rigid Structure-from-Motion: A Sequence-to-Sequence Translation Perspective}
%
%
%
%

\author{Hui Deng, Tong Zhang, Yuchao Dai, Jiawei Shi, Yiran Zhong, and Hongdong Li
  \IEEEcompsocitemizethanks{\IEEEcompsocthanksitem
    H. Deng, Y. Dai and J. Shi are with with School of Electronics and Information, Northwestern Polytechnical University and Shaanxi Key Laboratory of Information Acquisition and Processing, Xi'an, China.
    \IEEEcompsocthanksitem T. Zhang is with School of Computer and Communication Sciences, EPFL.
    \IEEEcompsocthanksitem Y. Zhong is with Shanghai AI Laboratory, Shanghai, China.
    \IEEEcompsocthanksitem H. Li is with School of Computing, the Australian National Universiy, Canberra, Australia.
    \IEEEcompsocthanksitem Y. Dai (daiyuchao@nwpu.edu.cn) is the corresponding author.}
}

%
%

\markboth{Journal of \LaTeX\ Class Files,~Vol.~14, No.~8, August~2015}%
{Shell \MakeLowercase{\textit{et al.}}: Bare Demo of IEEEtran.cls for Computer Society Journals}
%



\IEEEtitleabstractindextext{%
  \begin{abstract}
    Directly regressing the non-rigid shape and camera pose from the individual 2D frame is ill-suited to the Non-Rigid Structure-from-Motion (NRSfM) problem. This \emph{frame-by-frame} 3D reconstruction pipeline overlooks the inherent spatial-temporal nature of NRSfM, \ie, reconstructing the 3D sequence from the input 2D sequence.
    \REV{In this paper, we propose to solve deep sparse NRSfM from a \emph{sequence-to-sequence} translation perspective, where the input 2D keypoints sequence is taken as a whole to reconstruct the corresponding 3D keypoints sequence in a self-supervised manner.}
    \REV{First, we apply a shape-motion predictor on the input sequence to obtain an initial sequence of shapes and corresponding motions.}
    \REV{Then, we propose the Context Layer, which enables the deep learning framework to effectively impose overall constraints on sequences based on the structural characteristics of non-rigid sequences. 
    The Context Layer constructs modules for imposing the self-expressiveness regularity on non-rigid sequences with multi-head attention (MHA) as the core, together with the use of temporal encoding, both of which act simultaneously to constitute constraints on non-rigid sequences in the deep framework.}
    Experimental results across different datasets such as Human3.6M, CMU Mocap, and InterHand prove the superiority of our framework. The code will be made publicly available.
  \end{abstract}

  }

\maketitle

\IEEEdisplaynontitleabstractindextext

%
\IEEEpeerreviewmaketitle

\IEEEraisesectionheading{\section{Introduction}\label{sec:introduction}}
\IEEEPARstart{N}{on}-rigid structure-from-motion (NRSfM) aims at reconstructing the 3D shape of deforming objects from multiple 2D images, which is a central topic in geometric computer vision. It is well-known that NRSfM is an under-constrained (ill-posed) problem, thus various constraints on camera motions and non-rigid shapes have been introduced to provide regularization for the problem, such as shape basis (\emph{e.g.} single subspace \cite{bregler2000recovering}, union-of-subspace \cite{zhu2014complex}), trajectory basis \cite{akhter2008nonrigid}, shape-trajectory \cite{simon2016kronecker}, and temporal smoothness priors \cite{gotardo2011computing}.
Recently, deep neural networks have been applied to the NRSfM task and have shown an improved 3D reconstruction performance in terms of both accuracy and inference speed~\cite{cha2019unsupervised,novotny2019c3dpo,kong2020deep}.

However, most of these existing deep NRSfM methods perform non-rigid shape reconstruction from 2D observations in a \emph{frame-by-frame} manner, overlooking the aforementioned \emph{whole sequence} constraints that have been exploited by traditional methods for decades. \REV{As a result, the performance of current methods may be limited due to a potential lack of utilization of the sequence structural information.}

\begin{figure}[htbp]
  \centering
  \includegraphics[width=0.95\linewidth]{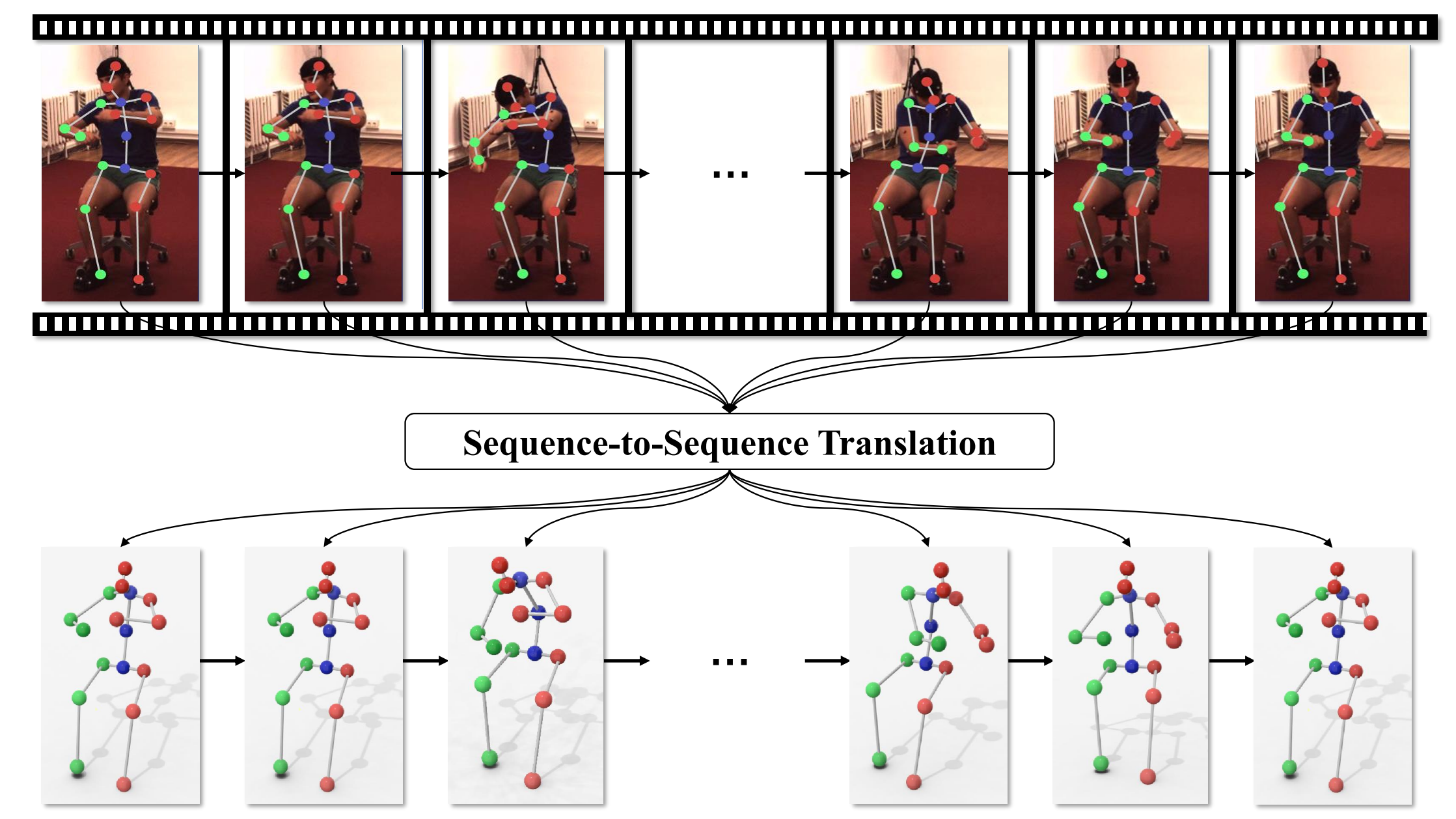}
  \caption{A conceptual illustration of our sequence-to-sequence NRSfM reconstruction framework, where we take the 2D frame sequence as a whole to predict the deforming 3D shape sequence.}
  \label{fig:head}
\end{figure}

There are two kinds of properties/constraints that are widely used in the traditional NRSfM methods: temporal smoothness of the deforming shape
sequence~\cite{dai2017dense,agudo2018robust} and low rank constraint of complex non-rigid shapes~\cite{dai2014simple,zhu2014complex}.
However, applying these constraints to the whole sequence in the deep neural networks framework is not straightforward.
The main challenges are: \textbf{1}) The existing single frame-based non-rigid reconstruction methods are hampered by the fact that expressing the whole sequence restrictions necessitates accepting the entire sequence as input; \textbf{2}) The motion ambiguity undermines the non-rigid shapes' representation;
\textbf{3}) Traditional non-rigid 3D shape sequence representations such as union-of-subspace representation~\cite{zhu2014complex} are difficult to fit the end-to-end training framework.

Inspired by the sequence-to-sequence model in machine translation, we propose to solve NRSfM from a sequence-to-sequence translation perspective, where a 2D sequence is ``translated'' into a 3D sequence. We assume that continuous 3D non-rigid deformation shape sequence has particular property. From this perspective, our \emph{sequence-to-sequence} reconstruction framework takes the input 2D sequence \textbf{as a whole} to reconstruct the deforming 3D non-rigid shape sequence. First, to minimize the motion ambiguity, a shape-motion predictor is applied to estimate the initial 3D shape and camera motion from every single frame to obtain coarse 3D shape sequence.
\REV{Then, we propose a context modeling module that models a 3D shape sequence by exploiting the inherent structure within the whole sequence, which aims to enforce the self-expressiveness  property}.
In addition, our proposed structure has sequence length scalability. The network is able to process longer input sequences at inference time, while keeping the length of the input sequences in the training phase constant.



\REV{We conduct experiments on various types of deformable objects, such as the human body datasets Human3.6M and CMU Mocap, as well as the hand dataset InterHand2.6M.}
Our main contributions are summarized as below:
\begin{compactitem}
  \item[$\bullet$] We propose a self-supervised sequence-to-sequence translation perspective to deep NRSfM, where the input sequence is taken as a whole to reconstruct the 3D non-rigid shapes. This is a paradigm shift from the current deep NRSfM pipeline.
  \item[$\bullet$] \REV{We propose a new method for introducing the self-expressiveness properties to sequence data, using multi-head attention as the core constructed module. In contrast to prior research on self-expressiveness networks, our method can achieve self-supervised training and is scalable in terms of sequence length. }
  \item[$\bullet$] Our method achieves state-of-the-art performance across multiple benchmarks compared to traditional and deep NRSfM methods, showing the superiority of our method in dealing with sequential data.
\end{compactitem}

\section{Related Work}
Here, we briefly review traditional and deep NRSfM methods, and related work in sequence modeling.

\subsection{Traditional NRSfM Methods}
Aiming at recovering both the deforming 3D shapes and camera motion from 2D measurements, non-rigid structure-from-motion (NRSfM) is highly ill-posed. Thus various constraints either on the camera motions or on the 3D shapes have been introduced to regularize the problem. In 2000, Bregler \etal \cite{bregler2000recovering} proposed to represent the deforming 3D shapes as a linear combination of a small number of basis shapes, \ie, a low-rank subspace representation. Later, Xiao \etal \cite{xiao2004closed} showed that there is further ambiguity in the shape basis representation. Akther \etal \cite{akhter2008nonrigid} introduced the DCT bases in the trajectory space. Torresani \etal \cite{torresani2008nonrigid} added a constraint on the shape distribution by introducing hierarchical prior. Simon \etal \cite{simon2016kronecker} unified shape and trajectory models by introducing the Kronecker-Markov prior.

Dai \etal \cite{dai2014simple} enforced the low-rank prior on the reshuffled shape representation, thus achieving prior-free reconstruction. Although this work can effectively solve the 3D structure and camera motion at the same time, the ambiguity problem between the camera pose and the rigid motion of objects still cannot be solved. Lee \etal~\cite{lee2013procrustean} proposed to solve this ambiguity by Procrustean Analysis, they defined the Procrustean Normalized Distribution to model 3D structure independently, without relying on the solution of camera motion.
Besides the above single subspace based representation, the union-of-subspace representation has been used to model complex non-rigid shapes \cite{zhu2014complex,agudo2020unsupervised}.

The methods discussed above focus on the reconstruction of sparse non-rigid objects, and there have been recent advances to achieve dense reconstruction by exploiting spatial constraints. Garg \etal \cite{garg2013dense} introduced the Total Variation norm into the model to constrain dense features. Kumar \etal \cite{kumar2016multi} performed subspace clustering in the trajectory space to achieve segmentation between different objects and organic priors~\cite{kumar2022organic} for non-rigid structure-from-motion. At present, many methods of processing multiple NRSfM are based on image segmentation~\cite{agudo2016recovering,kumar2016multi,kumar2017spatio}.
\REV{Although nonlinear models of non-rigid sequences are too complex to solve, and earlier models with a single subspace are too simplistic, union-of-subspace models strike a balance. 
Based on the model and solution theory of union-of-subspace, we have successfully extended the traditional sequence-to-sequence non-rigid 3D reconstruction setup to the deep framework.}

\subsection{Deep NRSfM Methods} Recently, deep models have been introduced to solve NRSfM \cite{bai2020deep,xu2021gtt,wang2021paul,sidhu2020neural}. Novotny \etal \cite{novotny2019c3dpo} proposed a concept of shape structure called transversal property, and designed a deep model combining canonicalization network and factorization network to rebuild camera motion and non-rigid shape. Kong \etal \cite{kong2020deep} used sparse dictionary learning by introducing block sparsity prior to building a multi-layer network structure to rebuild the 3D structure, which has also been extended to the perspective projection model to reconstruct the field scene \cite{wang2020deep}.\REV{However, the methods mentioned above mainly use a single-frame reconstruction approach. This means that they predict the corresponding 3D structure from a single frame of 2D observations. The lack of information from the sequence context makes it difficult to deal with the problem of endogenous ambiguity in NRSfM.} To eliminate the ambiguity between camera motion and non-rigid deformation, Park \etal \cite{park2020procrustean} used the Procrustean regression \cite{park2017procrustean} to model canonical shape \cite{park2020procrustean}. Recently, Zeng \etal \cite{zeng2021pr} proposed to use the Residual-Recursive Network and pairwise regularization to reconstruct and further regularize the reconstruction results. Wang \etal~\cite{wang2021paul} combined the neural network with the optimization algorithm, then recover camera pose and depth through bi-level optimization.

The above works focus on recovering the 3D keypoint coordinates of objects to achieve the sparse reconstruction of non-rigid objects. Recently, there have been approaches to reconstruct dense 3D shapes using differentiable rendering techniques\cite{yang2021lasr},\cite{yang2021viser},\cite{yang2022banmo}. Yang \etal~\cite{yang2021lasr} proposed a template-independent method for recovering the surface mesh of non-rigid objects from monocular video. Subsequently, embedding matching \cite{yang2021viser} and implicit neural expression \cite{yang2022banmo} were introduced into LASR\cite{yang2021lasr}, \REV{which achieves a more detailed surface reconstruction of the 3D model}.
\REV{This paper focuses on the structural properties of non-rigid sequences and implements a sequence-to-sequence reconstruction pipeline. A sparse setup is used for simplicity in the 3D representation.}

\subsection{Sequence Modeling}
Recurrent Neural Network (RNN) was first proposed by \cite{rumelhart1986learning},\cite{werbos1990backpropagation} for the sequence-to-sequence problem. RNN treats sequences as changes in state as time advances and effectively exploits the intra-sequence connections that are inaccessible to ordinary networks.
To deal with the gradient vanishing problem,  Long short-term memory(LSTM)~\cite{hochreiter1997long} and Gated recurrent unit(GRU)~\cite{chung2014empirical} make improvements upon RNN, and the modules enable the network to better cope with lengthened sequence by selectively retraining and forgetting information while processing important information in the sequence data more efficiently.
When combined with deep learning techniques, the neural network sequence model is better utilized to its advantage. Sutskever~\etal~\cite{seq2seq2014} first proposed the application of Deep Neural Networks(DNNs) with sequence modeling for Neural Machine Translation(NMT). Since then, such algorithmic design ideas have been quite successful in many sequence-to-sequence processing tasks, such as speech recognition~\cite{hinton2012deep} and image recognition~\cite{krizhevsky2017imagenet},\cite{he2017mask},\cite{redmon2018yolov3}. Compared with traditional sequence processing methods, a deep neural network based on LSTM~\cite{hochreiter1997long} module can understand the structure of the token sequence more effectively, and incorporate the sequence context information into the calculation organically.

One of the most significant achievements in recent years is the proposed transformer structures~\cite{vaswani2017attention}. In studying how to handle sequence context, researchers have further proposed autoregressive models~\cite{radford2018improving} and autoencoder models~\cite{devlin2018bert}, as well as a combination of them~\cite{lewis2020bart}. Its powerful understanding ability in long sequence processing makes it quickly overtake recurrent neural networks as the main choice in Natural Language Processing\cite{Beltagy2020Longformer}\cite{qin2022cosformer}. Then sequence modeling structure has been widely applied to various fileds, such as speech recognition~\cite{baevski2020wav2vec}, video captioning~\cite{Zhang2020ObjectRG}, action recognition~\cite{choi2019sdn}, human body pose estimation~\cite{kocabas2020vibe} and \emph{etc}.

Also, some work try to obtain contextual information without relying on these special structures. Park~\etal~\cite{park2020procrustean} proposed an optimization pipeline for reconstructing the corresponding 3D structure from a 2D keypoints set with the help of multi-layer perceptron and its numerical derivation and optimization algorithm, using Procrustean analysis as the core. Kanazawa~\etal~\cite{kanazawa2019learning} proposed a \textit{hallucinator} to make the network with the ability to indirectly exploit the temporal context information by knowledge distillation. Zeng~\etal~\cite{zeng2021pr} modeled the sequences from a unique perspective by giving constraints on the internal structure.



\subsection{Uniqueness of Our Contribution} Different from existing deep NRSfM methods, we solve NRSfM from a new sequence-to-sequence translation perspective, where the input sequence is taken as a whole to reconstruct the deforming 3D non-rigid shapes in a self-supervised learning manner. Thus, existing domain knowledge in modeling NRSfM such as union-of-subspace representation, rank minimization \etc could be incorporated into the deep framework naturally, which is a paradigm shift from the current frame-to-frame deep NRSfM pipeline.

\section{Method}
In this section, we present our solution to deep NRSfM. We start by recapping the commonly used constraints in NRSfM and then show how we implement them in our sequence-to-sequence translation framework. A theoretical proof is also provided to prove the validity of our design.

\begin{figure*}
  \centering
  \includegraphics[width=0.98\linewidth]{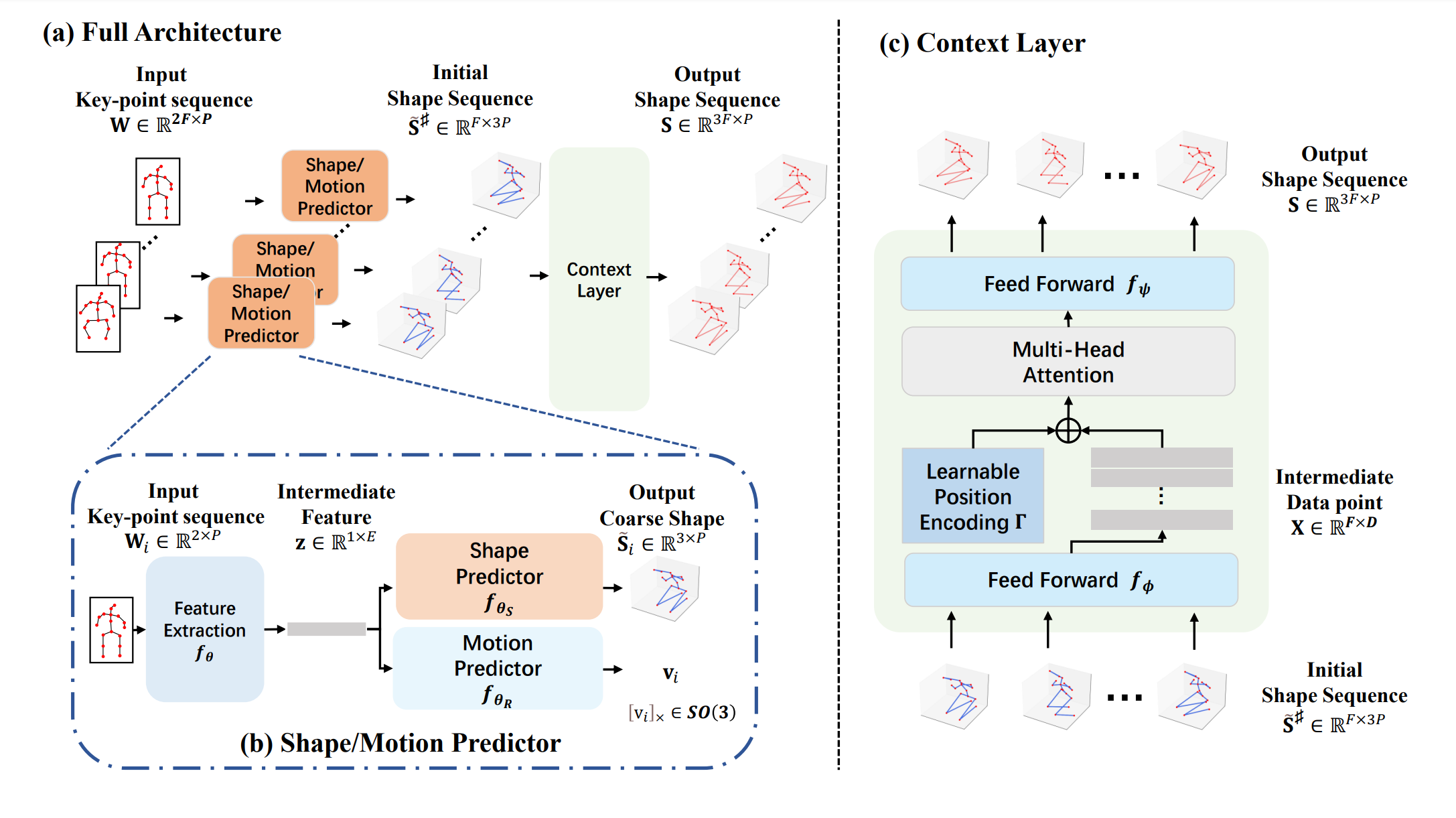}
  \caption{An overview of our proposed deep sequence-to-sequence NRSfM framework. Our framework consists of two core modules: Shape/Motion predictor (b) for estimating the initial 3D shape and camera motion from a single frame, and Context Layer (c) for adjusting the 3D shape sequence by exploiting the inherent structure within the whole 3D sequence.}
  \label{fig:overview}
\end{figure*}





\subsection{Problem Formulation and A Recap}
Given 2D measurements $\mathbf{W}_i \in \mathbb{R}^{2\times P}$, NRSfM aims at recovering the 3D deformable shape $\mathbf{S}_i \in \mathbb{R}^{3\times P}$ and the corresponding camera motion $\mathbf{R}_i \in \mathbb{R}^{2\times 3}$ such that $\mathbf{W}_i = \mathbf{R}_i \mathbf{S}_i$. Specifically, let's stack all the $F$ frames of 2D measurements and all the $P$ points in a matrix form, we reach:
\begin{equation}
  \label{eq:motion_shape}
  \mathbf{W}
  = \left[ \begin{array}{c}
      \mathbf{R}_1 \mathbf{S}_1 \\
      \vdots                    \\
      \mathbf{R}_F \mathbf{S}_F \\
    \end{array}
    \right]  \\
  = \left[
    \begin{array}{ccc}
      \mathbf{R}_1 &        &              \\
                   & \ddots &              \\
                   &        & \mathbf{R}_F \\
    \end{array}
    \right]\left[
    \begin{array}{c}
      \mathbf{S}_1 \\
      \vdots       \\
      \mathbf{S}_F \\
    \end{array}
    \right]
  = \mathbf{R} \mathbf{S},
\end{equation}

where $\mathbf{R} = \mathrm{blkdiag}(\mathbf{R}_1,\cdots,\mathbf{R}_F) \in \mathbb{R}^{2F\times 3F}$ expresses the camera rotation matrix. To regularize the problem, various constraints have been introduced to enforce the non-rigid shapes $\mathbf{S}_i$ and camera motion $\mathbf{R}_i$. In Bregler's seminal work \cite{bregler2000recovering}, each non-rigid shape is represented in a single subspace, \ie, $\mathbf{S}_i = \sum^{K}_{j=1}c_{ij}\mathbf{B}_j$. This single subspace representation could be naturally relaxed as minimizing the rank of the recovered shape $\mathbf{S}$.

In \cite{dai2014simple}, Dai \etal ~introduced the reshuffled shape representation $\mathbf{S}^\sharp \in \mathbb{R}^{F\times 3P}$, which connects to $\mathbf{S}$ as
$\mathbf{S}^{\sharp}=g(\mathbf{S}) = [\mathbf{P}_X ~~\mathbf{P}_Y ~~\mathbf{P}_Z](\mathbf{I}_3\otimes\mathbf{S}),
$
where $g(\cdot)$ denotes the reshuffle operation to the nonrigid shape. $\mathbf{P}_X, \mathbf{P}_Y, \mathbf{P}_Z$ are row-selection matrices defined in \cite{dai2014simple}. Based on this operator, Dai \etal ~presented the block-matrix method and achieved superior results under the single subspace assumption.

Later the single subspace model has been extended to the union-of-subspace representation \cite{kumar2016multi,zhu2014complex}, where the shape resides in each subspace and could be expressed with shapes in the same subspace.
\REV{However, it is challenging to directly generalize the union-of-subspace model based on conventional optimization to the deep framework. Upon further investigation of its solution principle, we discover that this optimization process ultimately leads to the self-expressiveness of non-rigid sequences},
in particular, $\mathbf{S} = \mathbf{C}\mathbf{S}$. \REV{This means that any shape in a sequence with the self-expressiveness property can get itself by a weighted combination of the sequence as a whole. Based on this concept, a new sequence-to-sequence pipeline is constructed. }

\subsection{Framework Overview}
\label{rearrange_shape}
\REV{We first provide a comprehensive overview of the pipeline structure.} As illustrated in Fig.~\ref{fig:overview}, our sequence-to-sequence deep {\small NRSfM} framework consists of two modules: \textbf{1}) non-rigid shape and motion predictor; and \textbf{2}) context modeling to regularize the sequence. It takes the input 2D sequence as a whole to predict the 3D sequence output. First, we employ a single-frame predictor to generate the initial 3D shape sequence and camera motion from the 2D frame. 

Then, the non-rigid shapes are optimized within our proposed context modeling layer to impose the self-expressiveness property and temporal smoothness constraint. The entire pipeline is fully \textbf{self-supervised}, \ie, no requirements for the ground truth camera motion and the non-rigid shape. We introduce the strengths of traditional NRSfM into the deep framework, naturally realizing the transition from single-frame shape estimation to the sequence-to-sequence framework, \ie, learning a single frame shape and motion predictor, and enforcing the spatial-temporal constraints on the whole sequence.

\subsection{Shape and Motion Prediction}
One of the advantages of a deep neural network is that it can learn a good initial coarse 3D shape estimation from the training data. Given 2D input $\mathbf{W} \in \mathbb{R}^{2F \times P} $, we first encode the 2D shape by a feature extractor $f_{\theta_e}: \mathbb{R}^{2P} \rightarrow \mathbb{R}^{E}$ as shown in Fig.~\ref{fig:overview}. Then we design a shape predictor $f_{\theta_s}: \mathbb{R}^{E} \rightarrow \mathbb{R}^{3P}$ and a motion feature predictor $f_{\theta_R}:\mathbb{R}^{E} \rightarrow \mathbb{R}^{3}$ to estimate a coarse estimation of 3D shapes and recover the rigid motion for each input 2D frame.
The coarse reshuffled shape $\tilde{\mathbf{S}}^\#$ and its corresponding rotation $\mathbf{R}$ are predicted as:
\begin{align}
   & \mathbf{v_i} = f_{\theta_R}(f_{\theta_e}(\mathbf{W_i})), \label{eq:rotv}           \\
   & \tilde{\mathbf{S}}_\mathbf{i}^{\sharp} = f_{\theta_S}(f_{\theta_e}(\mathbf{W_i})).
\end{align}
To keep $\mathbf{R} \in \text{SO}(3)$, we apply the Rodrigues' rotation formula to the extracted vector $\mathbf{v}_i$ as: $\mathbf{R}_i=\exp([\mathbf{v}_i]_{\times})$, where $[\cdot]_{\times}$ denotes the cross-product.

\subsection{Sequence Modeling and Self-expressiveness}
\subsubsection{Modeling Complex Non-rigid Sequences.}
As analyzed in the classic NRSfM methods, complex nonrigid motion usually contains multiple primitives or simple actions, and every single shape in the sequence is always related to its neighbors.
\REV{Intuitively, this would make the sequence exhibit a particular structural feature. In particular, the work of Zhu~\etal~\cite{zhu2014complex} proposes that the optimal solution for non-rigid sequences can be viewed as a specific application of low-rank matrix row-space recovery. Based on this, a union-of-subspace model has been proposed using the low-rank representation (LRR)~\cite{liu2012robust} theory by them as follow:}
\begin{align}
\mathbf{S}=\mathop{\arg\min}_{\mathbf{S},\mathbf{C},\mathbf{E}}\quad\| & \mathbf{S}^{\sharp}\|_*+\|\mathbf{C}\|_*+\lambda\|\mathbf{E}\|_2,   \\
  \textit{s.t.} \quad                                                    & \mathbf{S}^{\sharp} = g(\mathbf{S}),                                \\
                                                                         & \mathbf{S}^{\sharp} = \mathbf{C}\mathbf{A}, \label{eq:self_express} \\
                                                                         & \mathbf{W} = \mathbf{R}\mathbf{S}+\mathbf{E},
\end{align}
where $\mathbf{S}$ represents the 3D shape sequence matrix, and $\mathbf{R}$, $\mathbf{W}$ represent the motion matrix, and the 2D observation matrix, $g(\cdot)$ is the reshuffled operator as defined in \cite{dai2014simple}, $\mathbf{A}$, $\mathbf{C}$ represent the dictionary matrix, and codebook matrix of the LRR model, \REV{$\mathbf E$ is the residual noise}.

Due to the under-constrained dictionary matrix, this model is hard to fit the end-to-end deep learning framework. Thus we could soften this model by replacing the dictionary matrix with a shape sequence matrix as $\mathbf{A}=\mathbf{S}$, such a trick is widely used in practical scenarios\cite {liu2012robust}.
\REV{The core step Eq.\eqref{eq:self_express} of union-of-subspace is transformed into the self-expressiveness of the sequence}, in particular, $\mathbf{S}^{\sharp}=\mathbf{C}\mathbf{S}^{\sharp}$. 
\REV{The discussion above has focused on different versions of the original optimization model. The main issue is how this model fits into the deep framework.}

However, little has been done in incorporating the concept of union-of-subspace/self-expressiveness into a deep learning framework. Firstly, prior research on self-expressiveness layer aims to reconstruct a signal itself by using a linear combination of other signals~\cite{ji2017deep}, which critically relies on the data fidelity term, namely the reconstruction error.
Secondly, the regularizer term, such as the $\ell_1$ or $\ell_2$ norm, needs to be tuned for different batches, which makes it hard to apply in batch training.


In this paper, a Context Layer is proposed to enforce the self-expressiveness constraints on the 3D non-rigid shape sequence
\begin{align}
  \mathbf{S} & = f_{\psi}(\text{MHA}(\mathbf{X})),      \\
  \mathbf{X} & = f_\mathbf{X}(\tilde{\mathbf S}^\sharp)
\end{align}
where $f_{\psi}:\mathbb{R}^{D} \rightarrow \mathbb{R}^{3P}$ is a residual block and $\text{MHA}$ denotes the multi-head attention module, $\mathbf{X}$ is the latent code of the initial 3D shape sequence, which will be mentioned in Eq.~\eqref{eq:latent_code} and Eq.~\eqref{eq:lantent_code2}.

\REV{The reason for building the Context Layer around MHA is that, inspired by \cite{zheng2021ham}, we notice a correlation between the MHA module and matrix optimization.}
To learn the self-expressiveness structure of the latent code sequence, the dictionary matrix is set as a linear combination of the sequence items. The code book represents the correlation between the latent code of each shape in the sequence, so it can be modeled by the auto-correlation matrix of the sequence. Then, the reconstructed latent code sequence can be denoted as follow:
\begin{align}
   & \tilde{\mathbf{X}}=\mathbf{C}\mathbf{A},                                                                                                                  \\
   & \mathbf{C} = \text{Softmax} \left(\frac{\mathbf{X} \mathbf{W}_i^Q \cdot \mathbf{W}_i^{K\top } \mathbf{X}^\top } {\sqrt{D}}\right) , \label{eq:scaled_dot} \\
   & \mathbf{A} = \mathbf{X} \mathbf{W}_i^V,
\end{align}
where $\mathbf{W}^K,\mathbf{W}^Q,\mathbf{W}^V$ are the fully connected layer for key, query, and value. \REV{The computation process uses the similarity degree of the members among sequences as the weight and reconstructs the original sequence to apply corrections. This process effectively adds the self-expressiveness characteristics to the original sequence.}

In addition, unlike the self-expressiveness layer\cite{ji2017deep}, this framework is capable of training in a completely self-supervised manner.
Thus, the self-expressiveness layer\cite{ji2017deep} is replaced by the softmax operator, which provides us sparse coefficients without constraints on the reconstruction error. Our recovered shapes will be obtained by performing one more feed-forward network after a linear combination.

Moreover, our proposed Context Layer module enforces the self-expressiveness constraints with a length-independent scaled-dot product, the Context Layer thus enables a more general understanding of the sequence structure. This structure of placing regular constraints in the forward process instead of loss improves the network's capability to understand the sequence structure. 
\REV{Therefore, in theory, our framework could be trained with a fixed length while inference with an arbitrary length sequence. In this way, this framework could ensure the training efficiency. However, the sequence also contains direction information, this could give us the data temporal constraint and we should also take temporal information into account.}


\subsubsection{Modeling Temporal Encoding}
The above self-expressiveness representations encode complex non-rigid sequences, giving the Context Layer the capability to understand the overall structure information of the sequence. However, this encoding fails to capture the essential temporal constraints in the sequence data.

Here, we have made attempts to combine temporal information into the calculation process. Each of these methods has different advantages and disadvantages in different use cases. In the following parts, a brief description of these methods will be given.



\textBF{Absolute sinusoidal temporal encoding}. Inspired by recent advances in the Transformers, such as BERT~\cite{devlin2018bert} and GPT~\cite{radford2018improving}, we design our temporal encoding to facilitate the network's comprehension of the chronological relationships in the sequence. Following the existing protocols, where they first encode the features of words or image patches, we encode our coarse 3D shapes by a feedforward network $f_{\Phi}: \mathbb{R}^{F\times 3P} \rightarrow \mathbb{R}^{F\times D}$ as shown in Fig.~\ref{fig:overview} (b), followed by adding the temporal encoding feature:
\begin{equation}
  \mathbf{X} = f_\mathbf{X}(\tilde{\mathbf S}^\sharp) = f_\Phi(\tilde{\mathbf{S}}^\sharp) + \mathbf{\Gamma},
  \label{eq:latent_code}
\end{equation}
where $\mathbf{\Gamma}$ is the temporal encoding vector, \REV{temporal information is added to the shape latent code in the form of bias.} $\mathbf{\tilde{S}^\sharp}$ is a reshuffle form of shape matrix which is mentioned in Eq.~\eqref{eq:self_express}. Although the design of positional encoding in NLP does not match the nature of 3D non-rigid shape sequence, \eg a series of shapes could appear in a revert order but it rarely happens in NLP, they have certain similarities, for example, neighboring shapes or words are almost always together. As a result, we begin our temporal encoding.

To ensure the flexibility of the temporal encoding during the training process, the temporal encoding is set as a learnable parameter. Also, to ensure that the position encoding converges toward characterizing the sequential information, the following form is used as the initial value:
\begin{align*}
  \mathbf{\Gamma}(t, 2i)   & = \mathrm{sin}(t/10000^{2i/D}),     \\
  \mathbf{\Gamma}(t, 2i+1) & = \mathrm{cos}(t/10000^{2i + 1/D}),
\end{align*}
where $t$ is the frame number, $i$ is the index of the latent vector, $D$ is the dimensionality of the feature vector, and $\mathbf{\Gamma}$ is a learnable parameter. It is worth stating that using this temporal encoding alone is insufficient to bring about a performance change in the network.

Although the learnable temporal encoding is initialized with sinusoidal encoding, it lacks effective constraints during the subsequent training. Therefore, this temporal encoding approach needs to be used in conjunction with \emph{timing-related downstream supervision} in order to achieve the corresponding effect. Such an approach gives the best estimation results for a fixed length of inference length, and to verify this we conduct relevant experiments later on.

\textBF{Relative temporal encoding}. The absolute position encoding treats all members of the sequence equally, but the fact is the more distant in time, the lower the correlation. Moreover, as mentioned above, the self-expressiveness property of the Context Layer enables the network to better understand the association of the members of the sequence data, allowing this structure to have a considerable degree of length scalability. This feature enhances the training efficiency and increases the practical scope of the framework.

However, because the parameter determination of the learnable absolute temporal encoding is highly correlated with the training strategy, \ie input length during training, the network with this encoding scheme has higher computational accuracy but loses the length scalability.

In addition to the design idea, relative temporal encoding differs from absolute temporal encoding in, that the encoding containing temporal information is not directly injected into the input tokens, but is involved in the computation process as part of the reconstruction of the token sequence as below:
\begin{align}
   & \mathbf{C} = \text{Softmax}\left(\frac{f_{e}(f_{\mathbf{X}}(\mathbf{\tilde{S}^\sharp}))}{\sqrt{D}}\right), \label{eq:lantent_code2} \\
   & f_\mathbf{X}(\tilde{\mathbf S}^\sharp) = \mathbf{X} .
\end{align}
To better depict the temporal structure of the sequence without losing the length scalability, we chose two excellent relative encoding schemes~\cite{su2021roformer}\cite{ofir2022alibi} as the basis and test different relative temporal encoding schemes separately. A brief description of the two encoding schemes is given below.

RoPE\cite{su2021roformer} encodes the relative position information in a rotation matrix and integrates the position information into the feature calculation process in the form of matrix multiplication.
\begin{align}
   & f_e(\mathbf{X}) = \mathbf{XW}^Q\mathbf{R_\Theta W}^{K\top}\mathbf{X}^{\top},
\end{align}
where $\mathbf{R_\Theta}$ is a rotation matrix, $\Theta$ is a pre-defined parameter of rotation degree, and the degree is related to the relative position of elements in the sequence data. ALIBI~\cite{ofir2022alibi} takes a different idea and adds a bias related to the relative position to the calculation process of attention score.
\begin{align}
   & f_e(\mathbf{X}) = \mathbf{XW}^Q\mathbf{W}^{K\top}\mathbf{X}^{\top} + m\mathbf{\Sigma},
\end{align}
where $\mathbf{\Sigma}\in \mathbb{R}^{F\times F}$ is a pre-defined position related matrix, that has fixed pattern and can be easily expanded for arbitrary input length, $m$ is a constant number that is related to the head number.

We test the encoding schemes mentioned above in the following section and analyze their performance.

\subsection{Shape Regularizer}
Instead of imposing the constraint on the latent code as subspace clustering~\cite{ji2017deep} \REV{in a fully supervised manner},  we \REV{only} apply the regularizer after decoding the latent code to 3D shapes, where the 3D shape sequence's properties can be fully utilized.


In this section, we will discuss various constraining terms for the network to generate desired 3D shapes, and prove the necessity of $\mathbf{S}^\sharp$ to represent a deformable shape sequence. Meanwhile, the reason why these regularizers are chosen will be discussed.

Ambiguity is one of the main difficulties of non-rigid 3D reconstruction. The rigid motion that cannot be decoupled between the object and the camera causes the ill-posedness. The ill-posedness is one of the sources of under-determination in traditional methods~\cite{xiao2004closed,brand2005direct}. 
\REV{This inherent ambiguity will introduce shape estimation errors.}

To reduce the loss of accuracy due to ill-posedness, various methods have been proposed. Park~\etal~\cite{park2020procrustean} proposed the use of Procrustean analysis to evade the ambiguity, Zeng~\etal~\cite{zeng2021pr} used pairwise alignment to exclude the loss of accuracy caused by the rigid transformation ambiguity. However, Park~\etal~\cite{park2020procrustean} did not enable an end-to-end inference process, while Zeng~\etal~\cite{zeng2021pr} had an excessive computational complexity, \REVX{this complexity results in the method's inability in handling longer input sequence. It also results in its inability to obtain more contextual information.}. By contrast, we propose to use a special representation $\mathbf{S}^\sharp\in\mathbb{R}^{F\times 3P}$ instead of $\mathbf{S}\in\mathbb{R}^{3F\times P}$ to organize the sequence data as follow, exploiting the low-rank properties of the special representation to achieve a reduction in ambiguity.

We consider that the shape sequences composed of inter-frame rigid transformation-free shape sequences obey the low-rank assumption. Dai~\etal~\cite{dai2014simple} first proposed to introduce the low-rank assumption into the traditional method, which elegantly solves the problem of underdetermined solution equations due to insufficient conditions. On this basis, we impose constraints on the shape estimation process by introducing the low-rank assumption as a supervisory term into the training process.
\begin{align}
  \mathbf{S}^\sharp = \arg\min_{\mathbf{S}^\sharp}\text{rank}(\mathbf{S}^\sharp).
\end{align}

\newtheorem{theorem}{Theorem}
\newtheorem{lemma}{Lemma}
\newtheorem{definition}{Definition}

However, the constraint on the rank is an NP-hard problem, and to make it solvable, we follow Dai~\etal\cite{dai2014simple} and relax this constraint to the nuclear norm constraint. Thus, for the global shape constraint, we penalize the nuclear norm of the reshuffled shape matrix $\mathbf{S}^\sharp \in \mathbb{R}^{F\times 3P}$ as the nuclear norm is known as the convex envelope of the rank function:
\begin{equation}
  \mathcal{L}_{\mathrm{norm}} = \|\mathbf{S}^\sharp\|_* \label{eq:nuclear},
\end{equation}
where $\|\cdot\|_*$ denote the nuclear norm. It is worth noting that we are using the special formation, reshuffled shape matrix $\mathbf{S}^\sharp$, rather than $\mathbf{S}$ as there exists an inherent per-frame rotation ambiguity in non-rigid 3D reconstruction, \ie,
$\mathbf{W}_i = \mathbf{R}_i \mathbf{S}_i = \mathbf{R}_i \mathbf{Q}_i \mathbf{Q}_i^{-1}\mathbf{S}_i, \mathbf{Q}_i \in \mathrm{SO}(3).
$

The low-rank condition does not change as $\mathrm{rank}(\mathbf{S}) = \mathrm{rank}(\mathbf{Q}^{-1} \mathbf{S})$, where $\mathbf{Q}=blkdiag(\mathbf{Q}_1,\cdots,\mathbf{Q}_F)\in\mathbb{R}^{3F\times3F}$. It shows the limitation of $\mathbf{S}$ in enforcing the low-rank constraint, \ie, it cannot distinguish the non-rigid shapes from their per-frame rotated version.


In \cite{dai2014simple}, Dai \etal ~presented the block-matrix method using the reshuffled shape $\mathbf{S}^{\sharp}$ and showed the superiority of this reshuffled shape representation over the original one. Here we analyze the role of the reshuffled shape in recovering low-rank structure.
\begin{theorem}
  \label{theorem_1}
  Suppose that a deformable shape $\mathbf{S}^\sharp \in \mathbb{R}^{F \times 3P}$ lies in a low-rank space, then we have:
  \begin{equation}
    \mathrm{rank}\left ( g\left (\mathbf{S}\right ) \right ) = \mathrm{rank} (\mathbf{S}^\sharp)\leqslant \mathrm{rank}\left (g\left (\mathbf{Q}^{-1}\mathbf{S}\right )\right ).
  \end{equation}
  Here, $\mathbf{Q} \in \mathbb{R}^{3F\times 3F}$ is the per-frame rotation ambiguity matrix defined as $blkdiag(\mathbf{Q}_{1},\cdots,\mathbf{Q}_{F}), \mathbf{Q}_{i} \in SO(3)$. The equal sign holds if and only if $\mathbf{Q}$ contains only one rotation component $\mathbf{Q}_1=\cdots=\mathbf{Q}_F$, \ie, a global rotation on all the frames.
\end{theorem}


We define $\mathbf{S}_{\ast}^{\sharp} = g(\mathbf{Q}^{-1}\mathbf{S})$, and suppose that there are only $s$ different rotation components in $\left \{ \mathbf{Q}_{i} \right \}_{i=1}^{F}$, which means any other matrix $\mathbf{Q}_{j}$ is the same as any of these $s$ components. Based on this assumption, we can conclude that $\min((s-1)K, 8K)<\mathrm{rank}(\mathbf{S}_{\ast}^{\sharp})\leqslant \min\left ( sK, 9K \right )$,~$1 \leqslant s \leqslant F$, where $K$ is the rank of 3D reshuffled shape matrix $\mathbf{S}^{\sharp}$.
To make the narrative flow better, we provide complete proof and in-depth analysis in the supplemental materials.

\newenvironment{thmbis}[1]
{\renewcommand{\thethm}{\ref{#1}}%
  \addtocounter{theorem}{-1}%
  \begin{theorem}}
    {\end{theorem}}




Theorem 1 states that the rotation ambiguity $\mathbf{Q}$ affects the rank of the reshuffled shape $\mathbf{S}^\sharp$ directly~\footnote{It is worth noting that the rotation ambiguity in Theorem 1 is different from the ambiguity in calculating the corrective transformation matrix \cite{brand2001morphable,brand2005direct,xiao2004closed,akhter2009defense}.}. If and only if the correct rotations are recovered (up to a global rotation), the rank of the reshuffled shape $\mathbf{S}^\sharp$ is minimized. Therefore, we can recover the unique low-rank deformable shape using the rank minimization approach. Within our proposed deep NRSfM framework, we employ the reshuffled shape representation.

Furthermore, as mentioned above, we propose self-expressiveness constraints on non-rigid shape sequences as well as low-rank constraints, both of which are based on the existence of non-rigid transformations of shapes only. Therefore, we require that there should be no rigid transformation of any two shapes in the sequence when the above constraint is applied. That is, for sequence $\mathcal{S}=\{\mathbf{S}_0, ..., \mathbf{S}_n\}$, any shape $\mathbf{S}_i\ \in\mathcal{S}$ in the sequence can obtain $\mathbf{S}_2\in\mathcal{S}$ by rigid transformation, if and only f. $\mathbf{S}_1=\mathbf{S}_2$.

To ensure there are no rigid motions among the shapes in sequence, we follow the \textit{Lemma 1} of C3dpo~\cite{novotny2019c3dpo} that a set of shapes have the property of transversal, if there exists a canonicalization mapping $\Psi:\mathbb{R}^{3\times P} \rightarrow \mathbb{R}^{3\times P}$ that could eliminate all rigid motion in shape $\mathbf{S}$, and the processed $\mathbf{S}$ is consistent with the original $\mathbf{S}$, thus for any random rotation $\mathbf{R}\in\mathrm{SO(3)}$, we always have $\mathbf{S} = \Psi(\mathbf{R}\mathbf{S})$. If shapes in the output shape sequence contain various rigid motions, there should be difference between original sequence and canonical sequence, so we add the canonical loss to ensure the transversal property of the estimation results:
\begin{equation}
  \mathcal{L}_{\mathrm{cano}} = \frac{1}{F\cdot{M}}\sum^F_i\sum^M_j{\|\mathbf{S}_{ij}-\Psi(\hat{\mathbf{R}}_j\mathbf{S}_{ij})\|_l},
\end{equation}
where $\Psi(\cdot)$ is the canonicalization network as C3dpo~\cite{novotny2019c3dpo}, and $\hat{\mathbf{R}}_j\in{SO(3)}$ is a randomly sampled rotation for $M$ times across $F$ frames.

As mentioned above, we also find out that, when the Context Layer reconstructs shape sequence with absolute temporal encoding, a timing-related downstream constraint is needed to help the network understand the role of temporal encoding during training. We chose a smooth constraint term as the downstream constraint to describe that, only minor non-rigid deformation between sequence data in a short period of time:
\begin{equation}
  \mathcal{L}_{\mathrm{smooth}} = \sum^{F}_{i=1}{\|\mathbf{S}_i-\mathbf{S}_{i-1}\|_2}.
\end{equation}
Also, we calculate the reprojection error between the estimation result $\mathbf{RS}$ and the input $\mathbf{W}$ to enforce the network to learn how to estimate the shape as well as the motion from the input:
\begin{equation}
  \mathcal{L}_{\mathrm{data}} = \frac{1}{F}\sum^{F}_i{\|\mathbf{W}_i-\mathbf{R}_i\mathbf{S}_i\|_l}.
\end{equation}
The total loss function is reached as:
\begin{equation}
  \mathcal{L} = \mathcal{L}_{\mathrm{data}} + \alpha\mathcal{L}_{\mathrm{norm}} + \beta\mathcal{L}_{\mathrm{smooth}} + \lambda\mathcal{L}_{\mathrm{cano}},
\end{equation}
where $\alpha$, $\beta$ and $\lambda$ are trade-off parameters. Our loss functions are all self-supervised, which share the same setting as traditional NRSfM methods.


\section{Experiments}
In this section, we compare our approach with both traditional NRSfM and deep learning-based NRSfM methods. 

\subsection{Datasets and Setups}

\subsubsection{CMU MOCAP.} For a fair comparison with existing methods, we follow the setup of \cite{zeng2021pr} and choose 9 subjects from 144 subjects of the CMU MOCAP datasest~\footnote{CMU Motion Capture Dataset available at \url{http://mocap.cs.cmu.edu/}}. We build a training set and testing set following \cite{kong2020deep}, where 80\% of the action sequences in each subject are used for training and 20\% are used for testing. To verify the robustness, all 3D shapes in the dataset were subjected to a random orthogonal projection to obtain 2D observations, while the coordinates of all 3D shapes in the sequence were centralized in the form of \cite{bregler2000recovering}. Thus the camera matrix to be estimated only contains pure rotations. To facilitate data input, we first split the motion sequence into multiple chunks according to the subject, and then during training, the data are reshuffled in each training round. For evaluation, we use the same criterion as \cite{zeng2021pr} and report the normalized mean 3D error on shapes.

\begin{table*}[t]
  \centering
  \tabcolsep=0.35cm
  \caption{Results on the long sequences of the \textbf{CMU} motion capture dataset. We follow the comparison in \cite{zeng2021pr}. Our result surpasses the state-of-the-art on both All and Unseen datasets which is not available during training. \textbf{DNRSFM} and \textbf{PR-RRN} train a model for each subject separately for testing while we train only one model for different subjects. Notice that many methods have significant gaps in performance on the unseen set versus the training set, while our method achieves consistent performance on both datasets.}
  \label{tab:cmu}
  \scalebox{1}{
    \begin{tabular}{c|c|ccccccccc|c}
      \hline
      \small
                              & Methods                        & S07            & S20            & S23            & S33            & S34            & S38            & S39            & S43            & S93            & Mean$\downarrow$            \\ \hline
      \multirow{5}{*}{All}    & CSF\cite{gotardo2011computing} & 1.231          & 1.164          & 1.238          & 1.156          & 1.165          & 1.188          & 1.172          & 1.267          & 1.117          & 1.189          \\
                              & URN\cite{cha2019unsupervised}  & 1.504          & 1.770          & 1.329          & 1.205          & 1.305          & 1.303          & 1.550          & 1.434          & 1.601          & 1.445          \\
                              & CNS\cite{cha2019reconstruct}   & 0.310          & 0.217          & 0.184          & 0.177          & 0.249          & 0.223          & 0.312          & 0.266          & 0.245          & 0.243          \\
                              & C3DPO\cite{novotny2019c3dpo}   & 0.226          & 0.235          & 0.342          & 0.357          & 0.354          & 0.391          & 0.189          & 0.351          & 0.246          & 0.299          \\
                              & \textbf{Ours}                  & \textBF{0.072} & \textBF{0.122} & \textBF{0.137} & \textBF{0.158} & \textBF{0.142} & \textBF{0.093} & \textBF{0.090} & \textBF{0.108} & \textBF{0.129} & \textBF{0.117} \\
      \hline
      \multirow{4}{*}{Unseen} & DNRSFM                         & 0.097          & 0.219          & 0.264          & 0.219          & 0.209          & 0.137          & 0.127          & 0.223          & 0.164          & 0.184          \\
                              & PR-RRN                         &
      \textBF{0.061}          &
      0.167                   &
      0.249                   &
      0.254                   &
      0.265                   &
      0.108                   &
      \textBF{0.028}          &
      \textBF{0.080}          &
      0.242                   &
      0.162                                                                                                                                                                                                                              \\
                              & C3DPO                          & 0.286          & 0.361          & 0.413          & 0.421          & 0.401          & 0.263          & 0.330          & 0.491          & 0.325          & 0.366          \\
                              & \textBF{Ours}                  &
      0.081                   &
      \textBF{0.139}          &
      \textBF{0.196}          &
      \textBF{0.191}          &
      \textBF{0.195}          &
      \textBF{0.097}          &
      0.089                   &
      0.139                   &
      \textBF{0.151}          &
      \textBF{0.142}                                                                                                                                                                                                                     \\ \hline
    \end{tabular}}
\end{table*}

\subsubsection{Human3.6M.} This is a large dataset containing various types of human motion sequences annotated with 3D ground truth extracted using motion capture systems \cite{h36m2014}. Two variants of the dataset are used: the first uses 2D keypoints by orthogonal projection of ground truth 3D keypoint for training and testing (Marked as GT-H36M), and the second uses 2D keypoints detected by HRNet \cite{Sun_2019_CVPR} ((Marked as HR-H36M)). For a fair comparison, we follow the evaluation protocol of \cite{kudo2018unsupervised} and evaluate the absolute error measured over 17 joints.

\begin{table}[t]
  \tabcolsep=0.1cm
  \caption{Experimental results on the Human3.6M datasets with ground truth 2D keypoint (Marked as GT-H36M) and HRNet detected 2D keypoints (Marked as HR-H36M) and InterHand2.6M (Marked as I26M). We report the mean per joint position error (MPJPE) over the set of test actions. As the source code of PAUL and PRN are unavailable at the time of testing, we mark the result as ``-''.}
  \centering
  \label{tab:nocmu}
  \begin{tabular}{c|cc|cc|cc}
    \hline
                                   & \multicolumn{2}{c}{GT-H36M} & \multicolumn{2}{|c|}{HR-H36M} & \multicolumn{2}{c}{I26M}                                               \\
    \hline
    Methods                        & MPJPE$\downarrow$                        & Stress$\downarrow$                         & MPJPE$\downarrow$                     & Stress$\downarrow$         & MPJPE$\downarrow$         & Stress$\downarrow$        \\
    \hline
    PRN\cite{park2020procrustean}  & 86.4                        & -                             & -                        & -             & -            & -            \\
    PAUL\cite{wang2021paul}        & 88.3                        & -                             & -                        & -             & -            & -            \\
    ITES\cite{xu2021invariant}     & 77.2                        & -                             & -                        & -             & -            & -            \\
    PoseDict\cite{xu2021invariant} & 85.7                        & -                             & -                        & -             & -            & -            \\
    C3dpo\cite{novotny2019c3dpo}   & 95.6                        & 41.5                          & 110.8                    & 56.3          & 9.8          & 6.2          \\
    DNRSfM\cite{kong2020deep}      & 109.9                       & 35.9                          & 121.4                    & 72.4          & 13.8         & 8.5          \\
    Ours                           & \textBF{72.5}               & \textBF{29.9}                 & \textBF{90.5}            & \textBF{36.7} & \textBF{8.9} & \textBF{6.1} \\
    \hline
  \end{tabular}

\end{table}

\subsubsection{InterHand2.6M} This is a large dataset containing various types of hand poses \cite{moon2020interhand2}. The 2D annotations in this dataset come from human annotators and a 2D keypoint detection network, which is different from the dataset using motion capture to obtain the ground truth of 2D key points. Single-hand pose motion sequence data (ROM data) are selected for our method evaluation. We collect all single-hand ROM data in the 5FPS H+M InterHand2.6M, where 80\% are used for training and 20\% are used for testing. We evaluate the absolute error measured over 21 joints.

\subsubsection{Implementation Details.}

\noindent\textbf{Architecture detail.} During the training process for all datasets, we set $\alpha=0.01$, $\beta=0.001$ and $\lambda=0.003$. For the shape-motion predictor, the feature extraction module $f_\theta$ consists of 6 fully connected residual layers with 1024 hidden layer dimensions and output dimensions, and the shape predictor contains 2 fully connected layers, where the input/output dimensions are 1024/10,10/$3P$. $P$ are the joint number of shape, it relates to the specific dataset. The motion predictor contains one fully connected layer, which has 1024 input dimensions and 3 output dimensions. In the Context Layer, the feedforward network $f_\phi$ is a linear layer with $3P$ input dimension and 408 output dimension and the head number of attention block is 8, while the feedforward network $f_\psi$ contains 4 fully connected layers with 1024 hidden layer dimensions. The network is trained with the Adam optimizer with an initial learning rate of 0.001 and a 10-fold decaying at 1800 update steps and 6400 update steps for the Human3.6M dataset, while the learning weight will decay at 3600 update steps and 9600 update steps for the CMU Mocap dataset. \REV{Unless otherwise stated in this section, the experimental architecture in Tables~\ref{tab:cmu}/\ref{tab:nocmu}/\ref{tab:cmu_sub} use absolute encoding paired with smoothness loss.}

\noindent\textbf{Training Setting} We will give details of our experimental setup in this section. \REV{To avoid excessive computational overhead, we divide the sequence dataset into multiple subsequences of the same length for training. We then perform sequence-to-sequence reconstruction on these subsequences. This raises the question of whether the length of the sequence during training should be the same as during inference. We found that the difference between training and inference length has a significant impact on performance. Additionally, the use of temporal encoding with length scalability also has a significant impact on performance, which we verify in details later in the experiments.}

\begin{figure*}[t]
  \centering
  \includegraphics[width=0.8\linewidth]{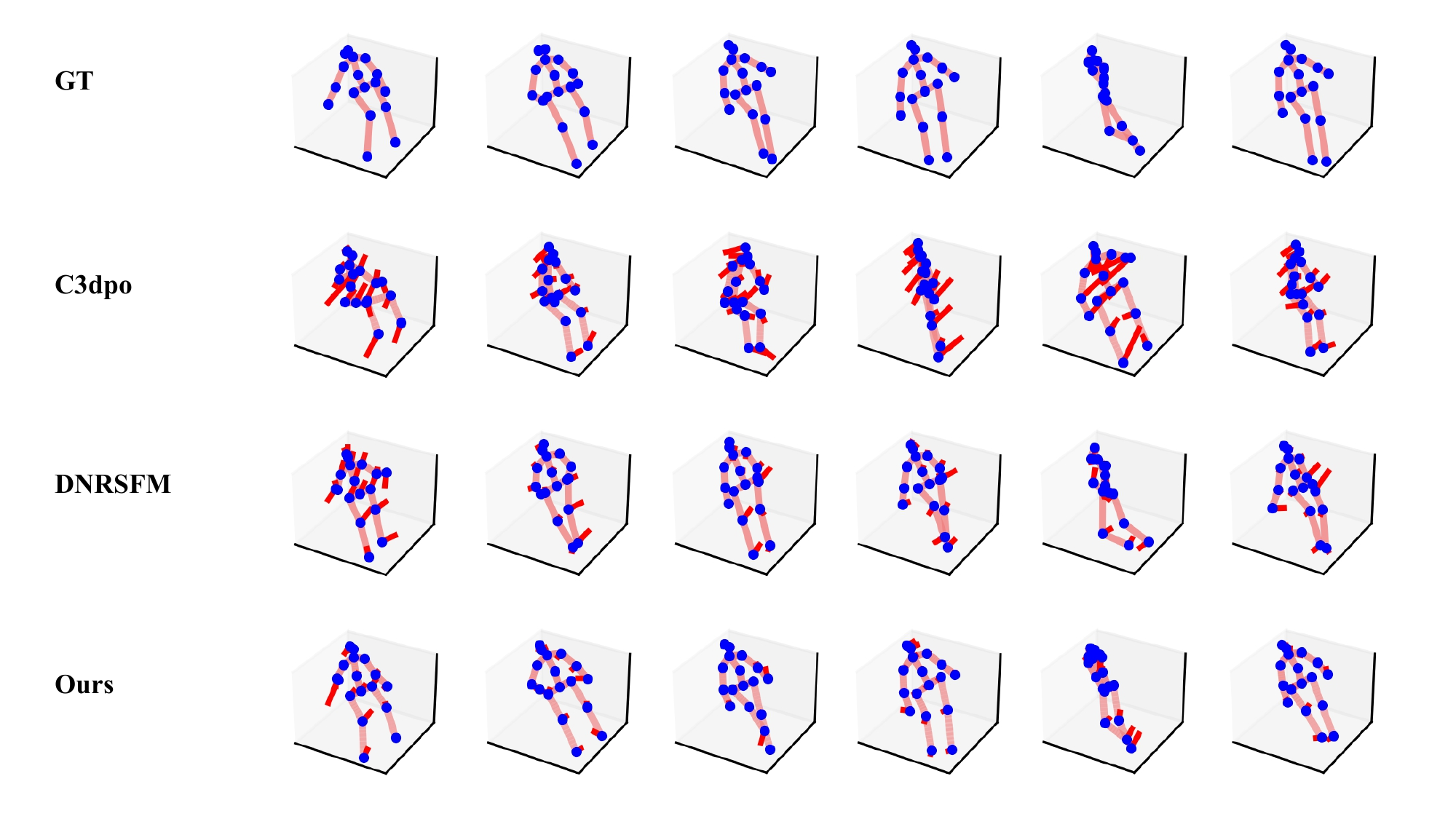}
  \caption{Qualitative comparison between our method and frame-to-shape method C3dpo \cite{novotny2019c3dpo} and DNRSfM \cite{kong2020deep} on Human3.6M, where we use the {\color{red}{red}} line to mark the error between the predicted 3D shape and the ground truth 3D shape.}
  \label{fig:result-c3dpo}
\end{figure*}


\subsection{NRSfM Results}
To guarantee fairness in performance evaluation, we follow the setting in \cite{wang2021paul} and utilize the same data preprocessing method as C3dpo\cite{novotny2019c3dpo}. We measure the performance of our method on the Human3.6M dataset by calculating the mean per joint position error(\textbf{MPJPE}), which is a widely used metric in the literature. To deal with depth flip ambiguity, we follow \cite{kudo2018unsupervised} to evaluate MPJPE for original prediction and depth-flipped prediction, and then retain the lower result.

In addition, we report the \textbf{Stress} metric following \cite{novotny2019c3dpo} on Human3.6M, which is invariant to the camera pose and the absolute depth ambiguity. Furthermore, we follow the setting in \cite{zeng2021pr} and utilize the same method as PR-RRN to preprocess the dataset. We use the normalized mean 3D error ($\textrm{e}_{3D}$) to measure the performance on the CMU MOCAP dataset. The MPJPE, Stress, and $\textrm{e}_{3D}$ metrics are calculated as:
\begin{align}
   & \textrm{MPJPE}(\mathbf{S}_i, \mathbf{S}^*_i) = \frac{1}{P}\sum_{j=1}^{P}{\|\mathbf{S}_{ij} - \mathbf{S}^*_{ij}\|},                                        \\
   & \textrm{Stress}(\mathbf{S}_i, \mathbf{S}^*_i)=\sum_{j<k}{\frac{\|\|\mathbf{S}_{ij}-\mathbf{S}_{ik}\|-\|\mathbf{S}^*_{ij}-\mathbf{S}^*_{ik}\|\|}{P(P-1)}}, \\
   & \textrm{e}_{3D}(\mathbf{S}_i, \mathbf{S}^*_i) = \frac{\|\mathbf{S}_i - \mathbf{S}^*_i\|_2}{\|\mathbf{S}^*_i\|_2},
\end{align}
where $\mathbf{S}_{ij}$, $\mathbf{S}^*_{ij}$ represent the $j$-th predicted and ground truth joint position coordinate of 3D shape on the $i$-th frame.

\begin{table}
  \centering
  \tabcolsep=0.87cm
  \caption{We report the average Normalized Error (NE) as \cite{wang2021paul} of various methods after training on a mixed-subject(subject 7, 20, 34 and 93) dataset of \textbf{CMU}.}
  \label{tab:cmu_sub}
  \begin{tabular}{c|cc}
    \hline
    Methods                      & Overall        & Unseen         \\ \hline
    C3dpo\cite{novotny2019c3dpo} & 0.371          & 0.364          \\
    DNRSFM\cite{kong2020deep}    & 0.201          & 0.219          \\ \hline
    \textbf{Ours}                & \textBF{0.191} & \textBF{0.203} \\ \hline
  \end{tabular}
  \vspace{-5mm}
\end{table}

\subsubsection{CMU MOCAP.} Experimental results on the CMU MOCAP dataset are reported in Table~\ref{tab:cmu}. We compare our method with competing methods. Since we used the same preprocessing method as PR-RRN\cite{zeng2021pr} which achieves state-of-the-art reconstruction accuracy, we directly cite the experiment results from them to compare with our experimental results. The mean reconstruction accuracy of our method on the \textbf{Unseen} set exceeds the results reported in PR-RRN and surpasses the state-of-the-art result in most of the subjects. Notice that we use a different training strategy from PR-RRN. PR-RRN trained different models for different shape subjects and tested them on their respective subjects, while we train only one model and test it on different subjects separately. Also, to ensure the completeness of the control experiments, we have followed our training strategy on a smaller subset CMU Mocap dataset for other methods as Table~\ref{tab:cmu_sub}, and the results demonstrate that our method still has advantages. It is worth mentioning that our method achieves similar performance on the unseen and training sets, while the performance of other methods is relatively different on these two datasets. 
\REVX{In summary, our method demonstrates a greater advantage in computational efficiency in comparison to PR-RRN, particularly given the high theoretical computational complexity of PR-RRN, which precludes it from taking longer sequences as input.}

\subsubsection{Human3.6M.} Table~\ref{tab:nocmu} reports results on Human3.6M with ground truth keypoint and results on Human3.6M with detected keypoints. Our method outperforms other methods by a large margin, and qualitative comparison is shown in Fig.~\ref{fig:result-c3dpo} and Fig.~\ref{fig:affinity}. We compare our result with several strong deep non-rigid reconstruction methods, where all but PRN~\cite{park2020procrustean} use single frame data as input to predict the corresponding 3D shape, and PRN does not impose temporal constraints in the computation. In contrast, our method explicitly models the sequence structure under the temporal constraint, taking advantage of the information ignored by other methods and thus achieving improved results. We visualize the estimation errors of our method and other methods in Fig.~\ref{fig:result-c3dpo}.

\subsubsection{InterHand2.6M.} 
\begin{figure*}[t]
  \centering
  \includegraphics[width=0.9\linewidth]{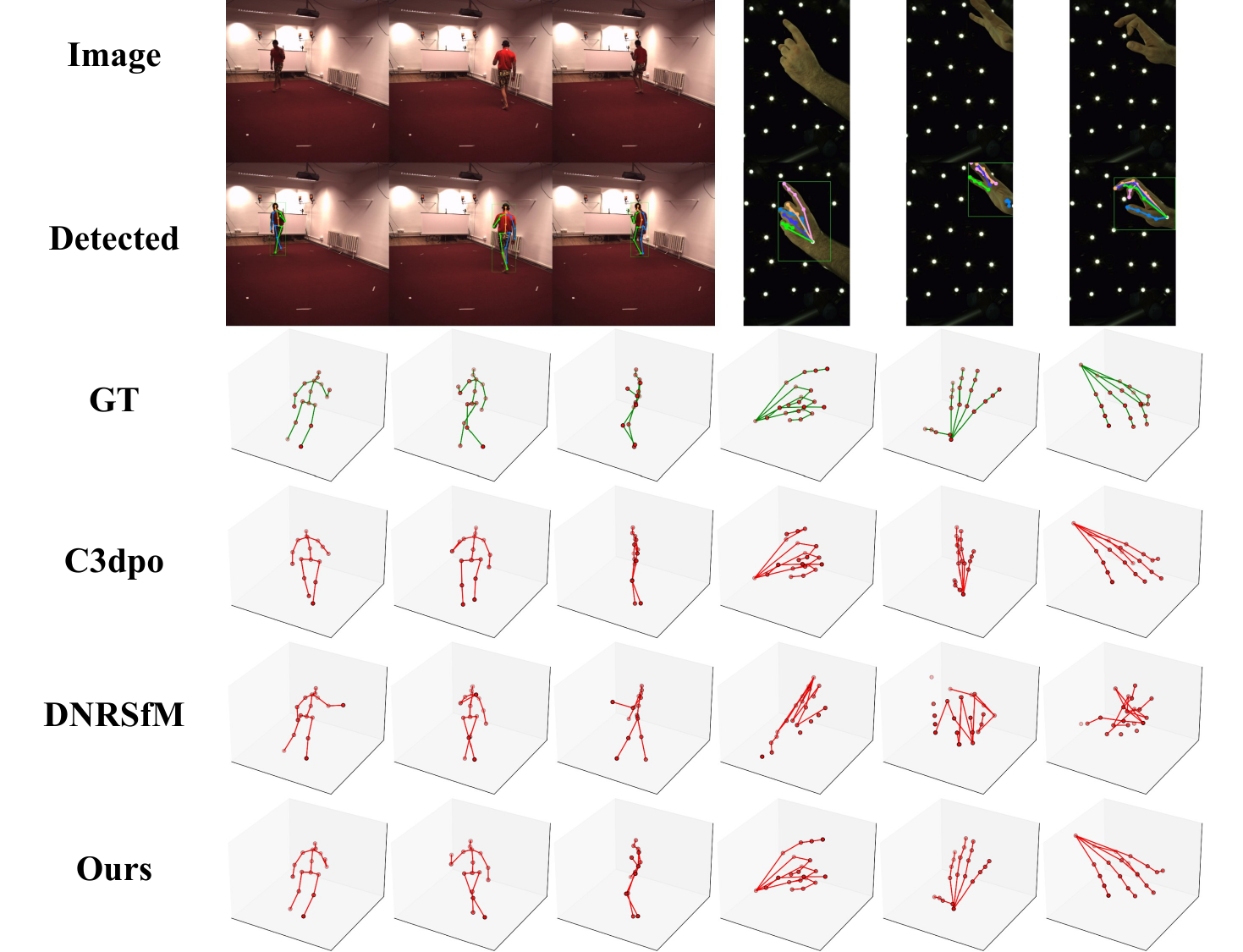}
  \caption{Visualization of several methods on Human3.6M with detected 2D keypoint and Interhand2.6M dataset. Our method is effective in reconstructing different kinds of non-rigid objects.}
  \label{fig:affinity}
\end{figure*}

We also compare our method with competing methods on the InterHand2.6M dataset. Quantitative results reported in Table~\ref{tab:nocmu} and the visualization results shown in Fig.~\ref{fig:affinity} show that our method achieves superior 3D reconstruction accuracy in a different category of NRSfM task.

\subsection{Model Analysis}
In this section, we conduct various ablation studies to validate the ability to model the sequence structure.

\begin{table*}[]
\caption{Qualitative comparison among different experiment settings and different input length on H36M, \textbf{Base+abspos} means a basic setting that including shape-motion predictor, Context Layer, learnable \textbf{absolute temporal encoding}, \textbf{Base+abspos+smooth} denotes basic setting with smoothness regularizer, \textbf{Base+nopos} denotes basic structure without temporal encoding, \textbf{RoPE, ALIBI} here means that the absolute temporal encoding is replaced with \textbf{RoPE} and \textbf{ALIBI}. (\textbf{Base+nopos+408linear, Base+nopos+512linear, Base+512linear+smooth, Base+512linear}) are trained with 32-length input sequence. The best results are labeled with \textBF{bold} font and second best results are \underline{underlined}. Because the sequence length does not exist for the single-frame methods, as a comparison their results appear only once in the table, with the rest marked by '-'.} 
\centering
\small
\label{tab:input_length}
\begin{tabular}{l|ccccccccc}
\toprule
                  \textbf{Inferernce Length}     & 32                   & 48                   & 64                   & 80                   & 96                   & 112                  & 128                  & 256                  & \multicolumn{1}{c}{512} \\ \hline
                       & MPJPE$\downarrow$                & MPJPE$\downarrow$                 & MPJPE$\downarrow$                 & MPJPE$\downarrow$                 & MPJPE$\downarrow$                 & MPJPE$\downarrow$                 & MPJPE$\downarrow$                 & MPJPE$\downarrow$                 & MPJPE$\downarrow$                     \\
\hline
\textbf{Training Length: 1}  & \multicolumn{1}{l}{} & \multicolumn{1}{l}{} & \multicolumn{1}{l}{} & \multicolumn{1}{l}{} & \multicolumn{1}{l}{} & \multicolumn{1}{l}{} & \multicolumn{1}{l}{} & \multicolumn{1}{l}{} & \multicolumn{1}{l}{}     \\ 
C3dpo                  & 95.6                 & -                    & -                    & -                    & -                    & -                    & -                    & -                    & -                        \\
DNRSfM                 & 109.9                & -                    & -                    & -                    & -                    & -                    & -                    & -                    & -                        \\
PAUL                   & 88.3                 & -                    & -                    & -                    & -                    & -                    & -                    & -                    & -                        \\ \hline
\textbf{Training Length: 32}  & \multicolumn{1}{l}{} & \multicolumn{1}{l}{} & \multicolumn{1}{l}{} & \multicolumn{1}{l}{} & \multicolumn{1}{l}{} & \multicolumn{1}{l}{} & \multicolumn{1}{l}{} & \multicolumn{1}{l}{} & \multicolumn{1}{l}{}     \\ 
Base+abspos            & 79.8                 & 162.5                & 168.3                & 174.5                & 177.5                & 181.7                & 184.5                & 199.2                & 204.8                    \\
Base+abspos+smooth     & \textBF{72.5}                 & 101.9                & 115.4                & 123.1                & 129.3                & 132.9                & 136.1                & 146.3                & 150.7                    \\
Base+nopos             & 78.3                 & \textBF{78.2}                 & \textBF{78.1 }                & \textBF{77.9}                 & \textBF{78.1}                 & \textBF{77.8}                 & \textBF{77.9}                 & \textBF{77.6}                 & \textBF{77.1}                     \\
Base+nopos+smooth      & 108.9                & 108.8                & 108.6                & 108.9                & 108.9                & 108.6                & 108.8                & 108.8                & 109.2                    \\
Base+rope              & 85.5                 & 88.2                 & 87.8                 & 88.2                 & 88.6                 & 89.1                 & 88.8                 & 89.9                 & 90.9                     \\
Base+rope+smooth       & 88.9                 & 105.9                & 116.7                & 124.3                & 129.2                & 134                  & 135.9                & 145.9                & 148.2                    \\
Base+alibi             & \underline{74.9}                 & \underline{80.2}                 & \underline{78.9}                 &\underline{79.3}                 & \underline{79.5}                 & \underline{79.6}                 & \underline{79.3}                 & \underline{79.2}                 & \underline{78.7}                     \\
Base+alibi+smooth      & 75.1                 & 81.9                 & 82.3                 & 81.3                 & 79.7                 & 81.2                 & 78.9                 & 80.3                 & 80.6                     \\ \hline
\textbf{Training Length: 256} & \multicolumn{1}{l}{} & \multicolumn{1}{l}{} & \multicolumn{1}{l}{} & \multicolumn{1}{l}{} & \multicolumn{1}{l}{} & \multicolumn{1}{l}{} & \multicolumn{1}{l}{} & \multicolumn{1}{l}{} & \multicolumn{1}{l}{}     \\ 
Base+nopos+408linear  & 101.9                & 103.1                & 103.6                & 104.4                & 105.5                & 106.3                & 107.6                & 110.8                & 113.8                    \\
Base+nopos+512linear         & 91.3                 & 91.5                 & 91.4                 & 91.9                 & 92.1                 & 92                   & 92.3                 & 93.5                 & 96.5                     \\
Base+512linear+smooth&97.6&103.1&106.9&108.6&111.8&113.7&116.4&129.3&154.3\\
Base+512linear & 98.4                 & 98.6                 & 142.6                & 99.3                 & 99.9                 & 100.2                & 100.4                & 103.3                & 119.2                    \\ \bottomrule
\end{tabular}
\end{table*}

\begin{figure*}[t]
  \centering
  \includegraphics[width=0.9\linewidth]{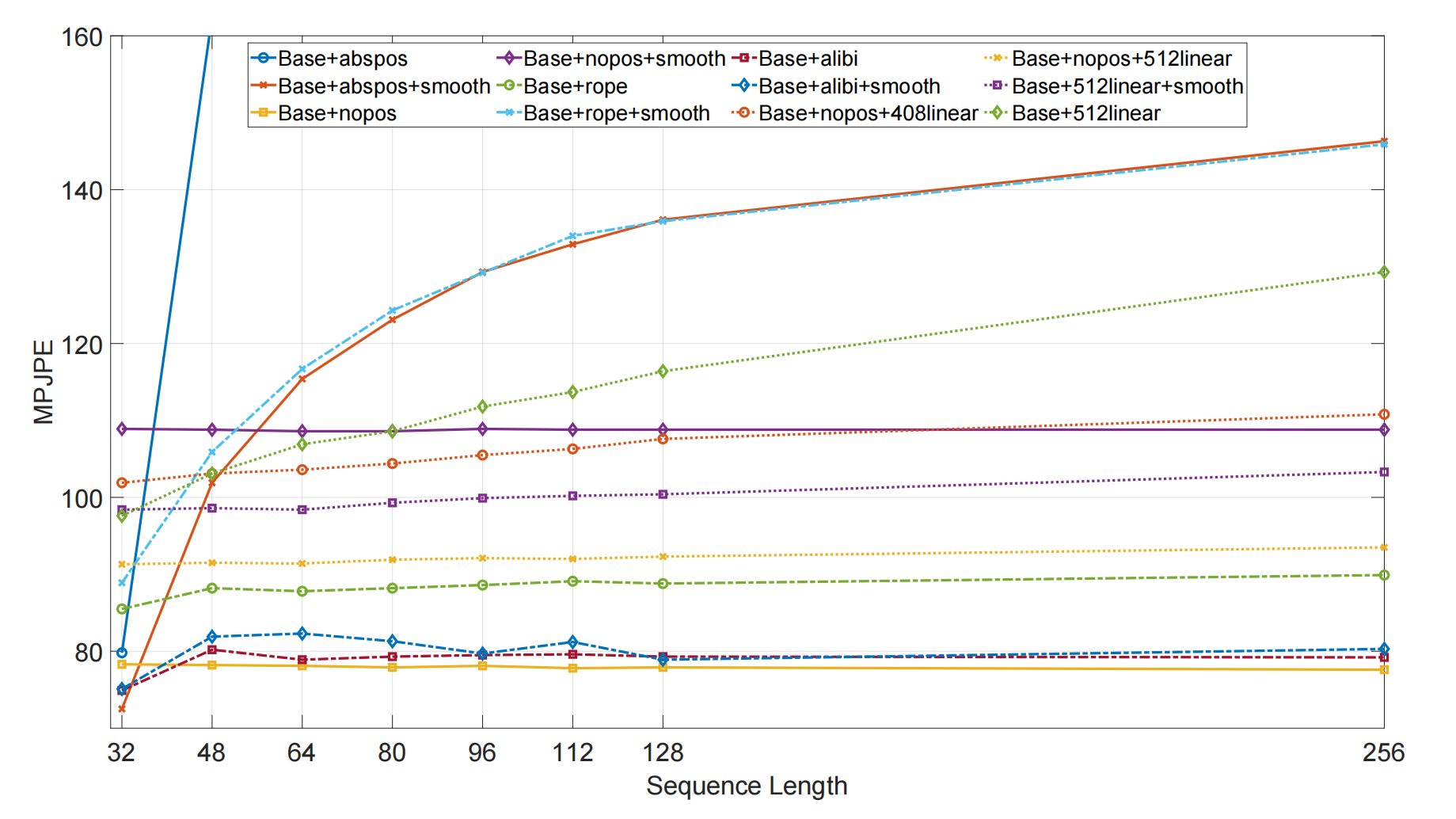}
  \caption{\REVX{Performance of each method with the increase in the length of the inference sequence for a certain training length.} The stability of the Base+nopos scheme in terms of length scalability and the optimal performance of the Base+abs+smooth scheme compared to other setups can be more intuitively seen in this figure.}
  \label{fig:input_length}
\end{figure*}

\subsubsection{Sequence Modeling}

 \REV{In this part, we discuss several key issues related to sequence modeling, including the performance difference between sequence and single-frame settings, the setting of regular constraints, and related ablation studies.}
 
\noindent\textbf{Whole sequence VS single frame:} To verify the importance of sequence modeling in non-rigid 3D reconstruction, we experimented by setting the input sequence length of our method to 1, which makes our method consistent with the frame-to-shape methods on experiment setup, and the result is reported in Table~\ref{tab:h36m-ablation} (labeled as \textbf{single frame}). After losing the sequence-structure information, our method suffers a significant performance loss, which demonstrates the contribution of sequence structural information in deep NRSfM.

\begin{table}[t]
  \centering
  \tabcolsep=0.4cm
  \small
  \caption{Ablation studies performed on the Human3.6M dataset}
  \label{tab:h36m-ablation}
  \begin{tabular}{l|ll}
    \hline
    Method                     & MPJPE$\downarrow$          & Stress$\downarrow$         \\
    \hline
    Ours(single frame)         & 90.7          & 39.4          \\
    Ours(shuffle)              & 83.7          & 36.6          \\
    Ours(reverse)              & 80.9          & 34.5          \\
    Ours(4-transformer layer)  & 93.6          & 40.6          \\
    Ours(w/o nuclear norm)     & 81.1          & 34.7          \\
    Ours(w/o canonicalization) & 98.1          & 40.7          \\
    Ours(w/o smoothness)       & 79.8          & 33.8          \\
    \textbf{Ours}              & \textbf{72.5} & \textbf{29.9} \\
    \hline
  \end{tabular}
\end{table}

\noindent\textbf{Low-rank constraint:} We also follow the low-rank assumption for sequences in traditional NRSfM~\cite{dai2014simple} modeling, thus we also design the nuclear norm term as Eq.~\eqref{eq:nuclear} to constrain the rank of the output shape sequence matrix. To verify its effectiveness, we design an ablation experiment without the nuclear norm term for training. As shown in Table~\ref{tab:h36m-ablation}, the unconstrained rank of the resulting sequence causes a performance decrease in our results.

\noindent\textbf{Canonicalization constraint:}
To avoid ambiguity in the estimation of non-rigid shapes, we follow C3dpo~\cite{novotny2019c3dpo} by incorporating canonicalization as a constraint such that no shape in the output sequence can be obtained from a rigid transformation of another shape, ensuring the uniqueness of the inference results. To verify the effectiveness of this constraint, we conducted ablation experiments, and the results are shown in Table~\ref{tab:h36m-ablation}. The canonicalization loss contributes significantly to the improvement of computational accuracy.

\noindent\textbf{Self-expressiveness:} To improve the training efficiency of the sequence-to-sequence model, we proposed a Context Layer that enforces self-expressiveness constraints as Eq.~\eqref{eq:self_express} in a length-independent manner. To verify the effectiveness of this module, we designed an ablation experiment with fixed training length and different inference lengths, and the results are shown in Fig.~\ref{fig:input_length}. Without retraining, the Context Layer can guarantee similar inference accuracy when processing different input lengths.




The self-expressiveness property shown in Eq.~\eqref{eq:self_express} indicates that the self-expressiveness of a sequence data is irrelevant to its length, and the Context Layer imposes self-expressiveness constraint on \REVX{predicted shape sequence} in a length-independent manner. Thus, if the Context Layer learns how to impose self-expressiveness constraints, a trained model should be able to impose self-expressiveness constraints on sequences of varying lengths correctly.
As shown in (\textbf{Base+nopos, Base+nopos+smooth, Base+nops+256}) Fig.~\ref{fig:input_length}, models trained with fixed-length sequences can cope with inference sequences of varying lengths and achieve similar reconstruction accuracy. This allows the network to be trained with shorter sequences no matter how long the expected inference sequence is. It significantly improves the network training efficiency. The problem of slow training speed \REVX{(0.7 hour per epoch in average with 32 sequence length, 60 epochs in total, more training speed details are provided in supplementary material)} and high computational cost caused by long training sequences is avoided.


\noindent\textbf{More attention blocks:}
To figure out whether more attention blocks contribute positively, we replace our Context Layer with a four-layer transformer. The results are reported in Table~\ref{tab:h36m-ablation}, which show that introducing more attention blocks has a negative impact instead. Our initial idea is to use attention blocks to achieve sequence modeling with self-expressiveness. However, as the experiments proved, only one attention block is already sufficient, while adding more blocks may lead to redundant parameters with insufficient constraints, which could make the network fail to converge in the desired direction and lead to the problem of performance degradation.

\subsubsection{Temporal Constraint}
\noindent\textbf{Sequence ordering:}
As a sequence-to-sequence deep NRSfM method, our method utilizes temporal information differently than PRN or PR-RRN. We introduce temporal information to allow the network to better perceive the structural information of the sequences. To verify this, we designed two comparisons: for a trained model, we input a set of disrupted order of the input sequences or reversed the order of the input sequences, and compared their respective results with those of the normal input. The results are shown in Table~\ref{tab:h36m-ablation} (labeled as \textbf{shuffle} and \textbf{reverse}). Reversing the input sequence does not change the continuity of the sequence, so it does not have much impact on the reconstruction accuracy, while completely disrupting the input order totally destroys the original continuity of the sequences, which has a drastic impact on the results.

\noindent\textbf{Temporal encoding:}
In this paper, we have proposed two temporal encoding methods to incorporate the temporal information into the feature vectors. To verify their respective performances, we designed experiments using different encoding schemes and the results are shown in Fig.~\ref{fig:input_length}. The absolute encoding with smoothness term achieves the best results when the inference length is the same as the training length. However, the parameters of the learnable absolute encoding are highly tied to the training length, resulting in a lack of length scalability for this encoding scheme. This drawback makes this scheme require a higher training overhead when coping with longer inference sequences.

As shown in Tab.~\ref{tab:input_length} and Fig.~\ref{fig:input_length}, the basic setting with smoothness term(\textbf{Base+abspos+smooth}) achieves a better result than others. On the other hand, the smoothness term performs poorly without absolute temporal encoding as results shown in \textbf{Base+nopos+smooth}, \textbf{Base+rope+smooth}, \textbf{Base+alibi+smooth}.

For the reason why the smoothing term does not work when paired with a structure other than the absolute position encoding, we have two different views for different cases.
\textbf{1)} The smoothing term is added to the loss function to help the network understand the role of position encoding to achieve better 3D shape sequence reconstruction.
When there is no encoding used in the Context Layer, the smoothness terms can mislead the learning direction of the network, resulting in worse training results.
\textbf{2)} The addition of smoothing term forces the network to focus on the local structure of the sequence, which has a similar effect to that of the relative temporal encoding, so that the smoothing term does not provide an effective aid to the structure using the relative position coding scheme.

Our framework can effectively explore the sequence information by combining temporal encoding with smoothing constraints. \REVX{In addition, the affect of sequence length variations on the effectiveness of the model is an issue worth investigating}. In the following, we will analyze it from two perspectives: the sequence length during training and the encoding scheme used in temporal modeling.


\noindent\textbf{Training sequence length:}
In order to figure out the possible effect of the training sequence length on the results, a set of comparison experiments using a training sequence of \textbf{256} length is prepared.
As the result shown in Fig.~\ref{fig:input_length}, \textbf{Base+nopos+408linear} means training with a 408-dimension shape decoder while \textbf{Base+nopos+512linear} uses a 512-dimension shape decoder. \textbf{Base+512linear} means training with a learnable absolute temporal encoding and \textbf{Base+512linear+smooth} means that an additional smoothing constraint is added to the previous setting. \REVX{The combined results indicate that, regardless of the additional adjustments made to the experimental setup, training with longer sequences is more challenging. Furthermore, longer training sequences do not result in superior experimental outcomes with the same training expenditure.}

\noindent\textbf{Encoding schemes:}
Fig.~\ref{fig:input_length} demonstrates the effect of using different encoding schemes on the experimental results. As shown in \textbf{Base+nopos}, the proposed module without adding any temporal encoding and smoothness constraints can obtain a fair experimental result with a strong length scalability.
And as shown in \textbf{Base+abspos}, \textbf{Base+abspos+smooth}, \textbf{Base+512linear} and \textbf{Base+512linear+smooth}, while the learnable temporal encoding scheme achieves better reconstruction results at a specific length, it reduces the length scalability of the module. Some relative temporal encoding methods\cite{su2021roformer}\cite{ofir2022alibi} are also tested.
Context Layer with RoPE\cite{su2021roformer}, as shown in Fig.~\ref{fig:input_length} (\textbf{Base+rope, Base+rope+smooth}), performs poorly when the length of inference sequence is inconsistent with training sequence. This is because RoPE is length-dependent context scheme during training. \REVX{In contrast, the scalable ALIBI\cite{ofir2022alibi} encoding scheme ensures that the encoding process is independent of the sequence length, thereby enhancing the scalability of the Context Layer while achieving more accurate estimation results.}

\begin{table*}[htbp]
\caption{Performance on H36M of different pipeline structures. We only show the best result for other methods and our original strategy. }
\centering
\small
\label{tab:more_test}
\begin{tabular}{l|cccccc}
\toprule
         Weight              & 0.1                   & 0.4                  &         1           & 1.2                   & 1.5                   & 1.8                  \\ \hline
                       & MPJPE$\downarrow$                & MPJPE$\downarrow$                 & MPJPE$\downarrow$                 & MPJPE$\downarrow$                 & MPJPE$\downarrow$                 & MPJPE$\downarrow$                     \\
\hline 
C3dpo                  &      -           & -                    & 95.6                    & -                    & -                    & -                                            \\
DNRSfM                 &  -               & -                    & 109.9                    & -                    & -                    & -                                            \\
PAUL                   &  -                & -                    & 88.3                    & -                    & -                    & -                                            \\ \hline
Independent Loss                 & 97.1                 & 119.4                & 112.5                & 89.8                 & 94.8                 & 94.5                                         \\
Frozen Predictor                  &  -                & -                    & 91.3               & -                    & -                    & -   \\
Origin    & -                 & -                & \color{red}\textbf{72.5}                & -                & -                & -                                    \\
\bottomrule
\end{tabular}
\vspace{-2mm}
\end{table*}

\subsubsection{Inference Sequence Length Analysis}
\REVX{Our proposed method takes the whole sequences as input rather than each single frame, which significantly differs from existing methods.} Here, we conduct experiments to study the robustness of our method to the input length during inference. To this end, we train our model with 32 frames as input length for 15 and 180 epochs, whereas, we use varying sequence lengths in the test stage to investigate the robustness under \textbf{Base+abspos+smooth} setting.

\begin{figure}[htbp]
  \centering
  \includegraphics[width=0.95\linewidth]{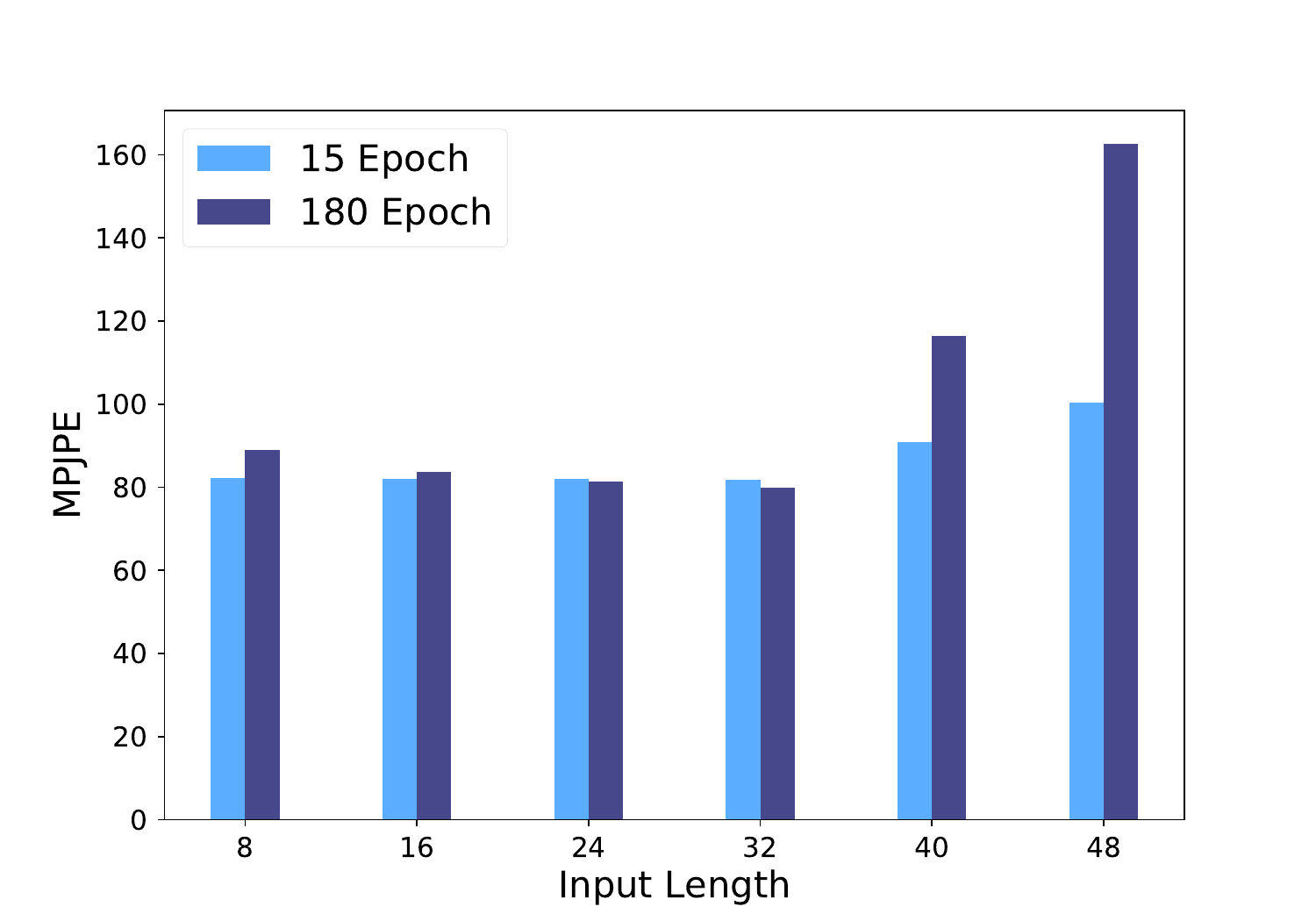}
  \caption{Experimental results on the Human3.6M dataset. We report the estimation accuracy for different input sequence lengths under the two models. }
  \label{fig:different_length}
\end{figure}

As shown in Fig.~\ref{fig:different_length}, a higher number of training epochs gives better results when the length of the input sequence at test time is the same as training.
Moreover, when the test sequence length is shorter than the training sequence, the test accuracy decreases gradually with the decrease of the sequence length. However, when the test sequence length is longer than the training sequence, we can observe the degradation of the test accuracy as the test sequence grows. It should also be noted that when the number of training rounds is relatively small, the accuracy degradation due to the change in input length is less dramatic compared with the case when the number of training epochs is large.

Regarding ALIBI~\cite{ofir2022alibi} and the above experimental results, we find that the model with learnable temporal encoding limits the length of the sequence that can be input during inference. Inference input length that differs from the training length introduces a loss of estimation accuracy. Moreover, when the number of training epochs is less, the network is less sensitive to changes in the input sequence length during inference. In this case, an inference input length different from the training length does not cause much accuracy loss. Conversely, a higher number of training epochs, although bringing higher estimation accuracy, also makes the network more sensitive to changes in the input sequence length. We believe this is due to the learnable temporal encoding gradually becomes associated with the training length during training, which in turn affects the module's ability to understand contextual information.

\subsubsection{Training Strategy Analysis}

\REVX{The role of the Shape/Motion Predictor is to estimate shape sequences from 2D observations. In order to ensure that the estimation results can obtain the self-expressiveness property, we jointly train the Shape/Motion Predictor with the Context Layer. Thus, we do not strictly require the Shape/Motion Predictor to produce \textbf{meaningful results at the beginning of the training}, where ``meaningful results'' refer to 3D sequences with sufficiently small re-projection error. During training, the Shape/Motion Predictor gradually converges to a certain direction, however, we do not require a 3D sequence with sufficiently small re-projection error.
}

\REVX{The reason for the above design is that we want the Context Layer to have more freedoms in reconstructing the 3D sequence. This will allow the Context Layer to act as a better regularizer, which leads to better results. Thus, constraint on freedoms of the initial sequences will not lead to better overall performance. Therefore, when the Shape/Motion predictor module is applied with geometric constraints or frozen after pre-training, the regularity of the Context Layer does not seem to work well with the Shape/Motion predictor. }

\REVX{To prove the point about freedom, we design the following ablation experiments on Human3.6M. The metric and training setting for these ablation is consistent with Table~\ref{tab:nocmu}.}

\begin{enumerate}
    \item \textbf{Independent Loss}. 
    \REVX{As mentioned above, we believe that the constraint on freedom of initial sequences will lead to a decrease in the overall performance. To prove this point, we add an independent re-projection loss for the Shape/Motion Predictor while preserving the other losses, and train it jointly with the Context layer. We set different loss weight as Table~\ref{tab:more_test}. It turns out that the training with this additional re-projection loss results in performance decrease in comparison to the original proposed method.}
    
    

    \item \textbf{Frozen Predictor}. 
    \REVX{To prove our point on freedom of initial output even further, we set up a more extreme strategy, where we first trained the Shape/Motion Predictor alone to the best it could be (with an MPJPE of 93 at H36M), followed by loading the weights in the pipeline and freezing them. The pipeline is then trained to analyze the effect of the Context Layer on the results.}
    \REVX{As shown in Table~\ref{tab:more_test}, although this strategy is able to obtain better results than the Independent Loss strategy, but not as good as original training strategy. Therefore, to obtain better results, \textbf{jointly training the Shape/Motion Predictor with Context Layer} is needed.}

\end{enumerate}

\REVX{To summarize, the ablation experiments show that our original proposed strategy gives the best results.}


\section{Conclusions}
In this paper, we have proposed a novel sequence-to-sequence translation perspective to deep NRSfM, where the input 2D sequence is taken as a whole to reconstruct the 3D deforming shape sequence. Our framework consists of a shape-motion predictor module and a context modeling module. We utilized a self-expressiveness mechanism and temporal position encoding to characterize complex non-rigid motion. Our method attempt to combine the strength of the traditional NRSfM and deep framework, \ie, the representation learning ability of neural networks and the sequence structure modeling from the conventional framework. Our framework is a paradigm shift from the current deep NRSfM pipeline. Experiment across different datasets (CMU Mocap, Human3.6M, and InterHand) demonstrate the superiority of our method over both traditional and deep NRSfM methods. In the future, we plan to extend our framework from sparse NRSfM to dense NRSfM.


%
%
%
%


%

%
%



\section*{Acknowledgment}
This work was partly supported by the National Natural Science Foundation of China (62271410) and the Fundamental Research Funds for the Central Universities.

\ifCLASSOPTIONcaptionsoff
  \newpage
\fi



%


\bibliographystyle{IEEEtran}
\bibliography{Deep_NRSfM_Reference}

%





\vspace{-10mm}
\begin{IEEEbiography}[{\includegraphics[width=1in,height=1.25in,clip,keepaspectratio]{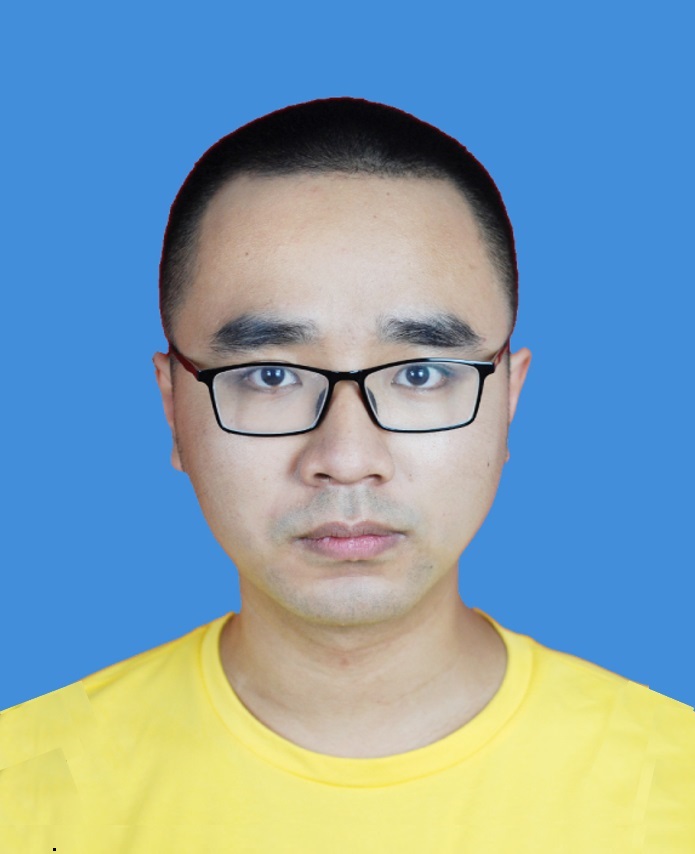}}]{Hui Deng} is a PhD. student with School of Electronics and Information at the Northwestern Polytechnical University (NPU), and his director is Professor Yuchao Dai. He received the B.E. degree, M.E degree in signal and information processing from NPU, Xi'an, China, in 2019 and 2022, respectively. His research interests include 3D vision, deep learning, optimization and rendering.
\end{IEEEbiography}

\begin{IEEEbiography}[{\includegraphics[width=1in,height=1.25in,clip,keepaspectratio]{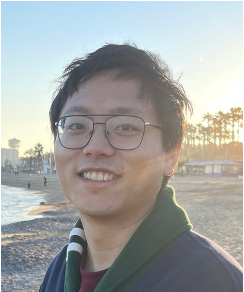}}]{Tong Zhang} received the B.S. and M.S degree from Beihang University, Beijing, China and New York University, New York, United States in 2011 and 2014 respectively, and he received the Ph.D. degree from the Australian National University, Canberra, Australia in 2020. He is working as a postdoctoral researcher at Image and Visual Representation Lab (IVRL), EPFL. He was awarded the ACCV 2016 Best Student Paper Honorable Mention and the CVPR 2020 Paper Award Nominee. His research interests include subspace clustering, deep geometric learning, and representation learning. \end{IEEEbiography}

\vspace{-12mm}
\begin{IEEEbiography}[{\includegraphics[width=1in,height=1.25in,clip,keepaspectratio]{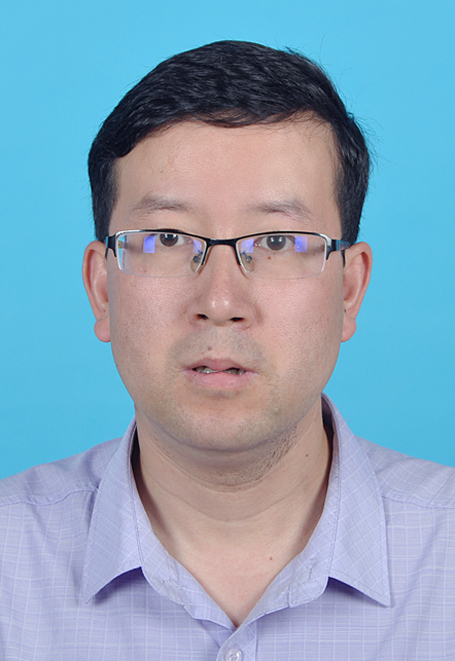}}]{Yuchao Dai} is currently a Professor with School of Electronics and Information at the Northwestern Polytechnical University (NPU). He received the B.E. degree, M.E degree and Ph.D. degree all in signal and information processing from NPU, Xi'an, China, in 2005, 2008 and 2012, respectively. He was an ARC DECRA Fellow with the Research School of Engineering at the Australian National University, Canberra, Australia. His research interests include 3D vision, multi-view geometry, low-level computer vision, deep learning  and optimization. He won the Best Paper Award in IEEE CVPR 2012, the Best Paper Award Nominee at IEEE CVPR 2020, the Best Algorithm Prize in NRSFM Challenge at CVPR 2017, the Best Deep/Machine Learning Paper Prize at APSIPA ASC 2017, etc. He served as Area Chair in CVPR, ICCV, ECCV, NeurIPS, etc.
\end{IEEEbiography}
\vspace{-12mm}
\begin{IEEEbiography}[{\includegraphics[width=1in,height=1.25in,clip,keepaspectratio]{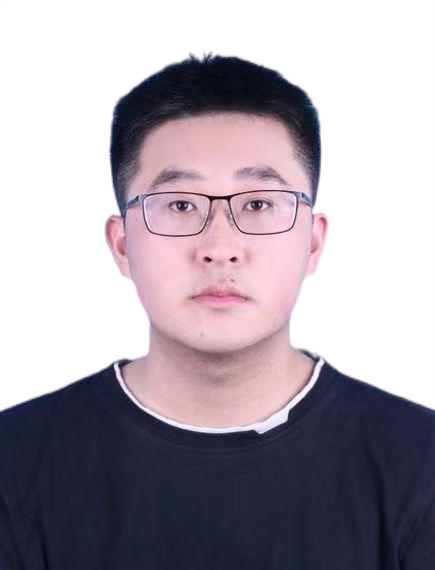}}]{Jiawei Shi} is a M.E student with School of Electronics and Information at the Northwestern
  Polytechnical University (NPU), and his director is professor Yuchao Dai. He received the B.E. degree in Information and Computational Science from NPU, Xi’an, China, in 2022. His research interests include computer vision and deep learning.
\end{IEEEbiography}
\vspace{-12mm}
\begin{IEEEbiography}[{\includegraphics[width=1in,height=1.25in,clip,keepaspectratio]{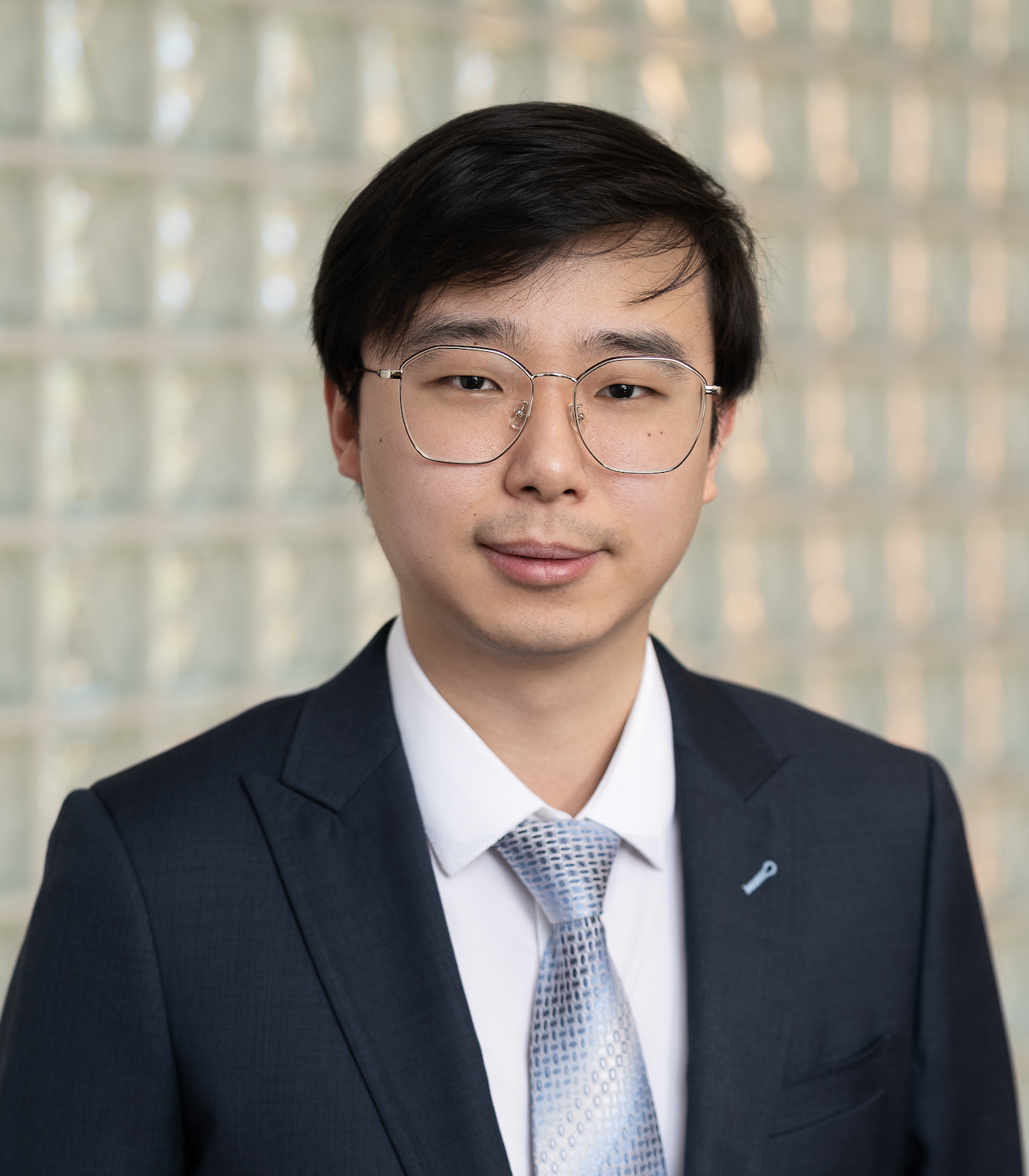}}]{Yiran Zhong} is a principal investigator at the Shanghai AI Laboratory, focusing on multimodal learning and autonomous driving. Prior to that, he received a Ph.D. degree in Engineering from The Australian National University, Canberra, Australia in 2021 and an M.Eng with the first class honor in engineering from The Australian National University, Canberra, Australia in 2014, and a B.E. degree from the University of Electronic Science and Technology of China in 2008. His research interests include self-supervised learning, visual geometry learning, multimodality learning, machine learning, and natural language processing. He won the ICIP Best Student Paper Award in 2014.
\end{IEEEbiography}

\vspace{-12mm}
\begin{IEEEbiography}[{\includegraphics[width=1in,height=1.25in,clip,keepaspectratio]{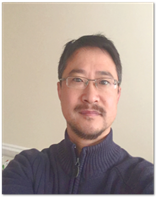}}]{Hongdong Li} is a Chief Investigator of the Australian Centre of Excellence for Robotic Vision, professor of Computer Science with the Australian National University. He joined the RSISE of ANU from 2004. He was a Visiting Professor with the Robotics Institutes, Carnegie Mellon University, doing a sabbatical in 2017-2018. His research interests include 3D computer vision, machine learning, autonomous driving, virtual and augmented reality, and mathematical optimization. He is an Associate Editor for IEEE Transactions on PAMI, and IVC, and served as Area Chair for recent CVPR, ICCV and ECCV. He was the winner of the 2012 CVPR Best Paper Award, the 2017 Marr Prize (Honorable Mention), a finalist for the CVPR 2020 best paper award, ICPR Best student paper award, ICIP Best student paper award, DSTO Fundamental Contribution to Image Processing Prize at DICTA 2014, Best algorithm award in CVPR NRSFM Challenge 2017, and a Best Practice Paper (honourable mention) award at WACV 2020. He is a Co-Program Chair for ACCV 2018 and Co-General Chair for ACCV 2022, Co-Publication Chair for IEEE ICCV 2019.
\end{IEEEbiography}




\end{document}